# From locomotion to cognition

## Table of contents

# 1 Summary of results (project period 1. 10. 2008 – 30. 9. 2009)

## (a) Work package 1: On-line manipulation of morphology

In our previous research, we have studied the role that animal and robot morphology plays in locomotor behavior and how this in turn influences higher-level cognitive capabilities. It turns out that to facilitate various functions and behaviors, morphologies are to be changed/tuned on-line. While this feature is omnipresent in animals, it has been largely missing in robots so far. In this work package, we are attempting to bridge this gap.

In various platforms, ranging from experimental platforms aimed at novel actuator development or focused studies of particular phenomena to complete running or swimming robots, we have been trying to address a number of issues. At the center of our investigations are passive joints. Joints that are passive (i.e. not actuated with a motor) provide a number of advantages. They require neither an appropriate control signal, nor energy, and can thus passively adapt to the interaction of the body with the environment. We explored several solutions. First, in underactuated platforms (swimming), we were investigating the distribution of the passive joints, i.e. which joints are best actuated and which passive. The passive joints themselves are typically not freely rotating, but passively compliant (they behave as a spring). The stiffness/compliance of every particular joint affects the behavior of the platform considerably. We have devoted time to explore such a stiffness distribution in joints off-line, but also approached the key challenge of how the robots can perform this change on-line. For that, we have also engaged ourselves in new actuator development. The joints turned to be of key importance, however, there are also other components of morphology that deserve attention. The first is shape: we have run tests with several shapes of trunks and legs (simulations of quadruped platforms) and body segments and fins (swimming platforms). The last part of morphology under investigation, which is important for control (WP 2 and 3), is the type and distribution of sensors.

While the previous paragraph outlines the topics we investigate, we have decided to structure this part of the report not according to those points, but according to the platforms we use. When describing these platforms and experiments, we will point out how this is relevant to the issues described above. This part is thus structured as follows: (i) Novel actuators, (ii) Legged platforms, (iii) Swimming platforms.

### Novel actuators

Passive joints with variable compliance (i.e. tunable online) are highly desirable to improve locomotor capabilities of robots, but devising satisfactory hardware solutions to this problem continues to be a challenge. In the past project period, we have developed two prototype solutions: (i) a magnetic spring, and (ii) a jack spring.

#### *Magnetic spring*

The “Magnetic spring” is an experimental platform designed and built by Emanuel Benker during his Bachelor's thesis work at our laboratory (Benker 2009).

**Concept.** When two magnets are placed in such a way that their north poles are facing each other, they repel with a force that depends on the inverse of the fourth power of the distance between the magnets. If we fix one magnet and the motion of the other one is constrained to the axis that goes through their middles, they can only approach or move apart from each other. The repulsion could be used as a restoration force and hence the system would resemble a compression non-linear spring. To control the stiffness of this nonlinear spring, we could use a coil to modify the magnetic field between the two magnets. A coil increasing the magnetic field would increase the repulsion; conversely, decreasing the field would decrease the repulsion. With these ideas in mind, we designed a magnetic spring and analyzed some of its properties.

**Position and length of the coil.** Before building the experimental prototype, we studied some parameters of the system by means of FEM simulations[1]. We focused on the position and length of the coil. In Figure 1, we show the results obtained with the simulations. The plot shows the force between the magnets at a nominal distance of 5 mm for different lengths and positions of the coil with a current fixed at 400mA. A zero position indicates that the outer extreme of the coil coincides with the outer extreme of the fixed magnet. The coil was assumed to have linear density of turns. From this result we choose an 8 mm coil that starts 2 mm behind the fixed magnet. The schematic of the experimental setup is also shown in Figure 1.

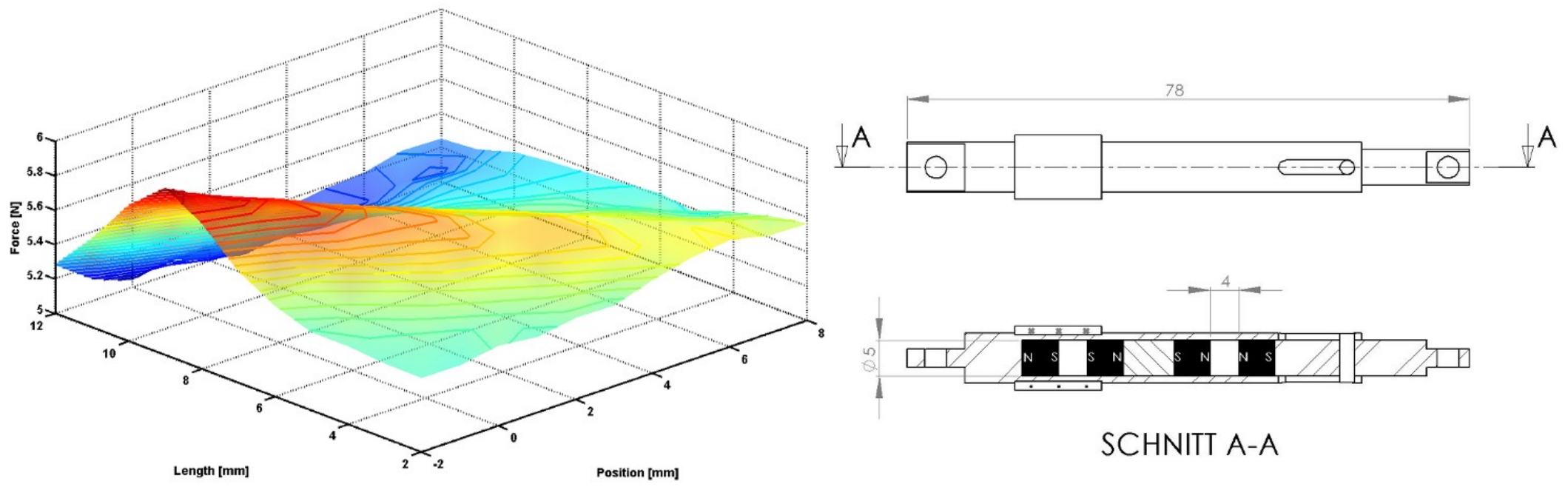

**Figure 1: Simulation of a magnetic spring.** On the left, force between the magnets at a nominal distance for different lengths and positions of the coil. Current at 400 mA. On the right, a schematic of the prototype. The fixed magnet lies inside the coil.

**Test platform.** Current through the coil was set and successively increasing values of force were applied using a dynamometer. A millimetric screw was used to measure the change in compression for each force value. The same procedure was repeated with decreasing forces; a slight difference is observed (not shown here) in the curves due to static friction.

**Spring curve modification.** To modify the curve of the spring, we need to supply different values of current to the coil. Two current values were used to observe maximum changes, namely -400 mA and 400mA. The results are shown in Figure 2. The overall peak to peak change in stiffness is approximately 10%.

**Advantages, drawbacks and future work.** The advantage of this setup is that it allows a real on-line adaptation of the spring curve with relative good force outputs (force output – weight ratio of about 60 N/Kg). On the other hand, the currents needed to produce an observable change are too high. The consequences of high currents are heat dissipation and high energy consumption.

Heat dissipation could finally destroy the magnets inside the system by taking them above the Curie temperature. A solution to this would be to improve the design by trying to reduce the current. A possible solution we are working on is to focus the magnetic field produced by the coil by means of a ferromagnetic core. Also, the geometrical arrangement is being modified into a 3-D structure rather than the axisymmetric one shown here. Given that to maintain a change in the spring curve, current has to be constantly supplied to the system, the consumption of a device like this is bound to be rather high. A work around is to store the change into a ferromagnetic material with high remanence, using current pulses, hence reducing the total energy needed. Though this idea is theoretically sound, we haven't succeeded in producing such a solution.

[1] D. C. Meeker, Finite Element Method Magnetics, Version 4.2 (15Jul2009 Mathematica Build), http://www.femm.info

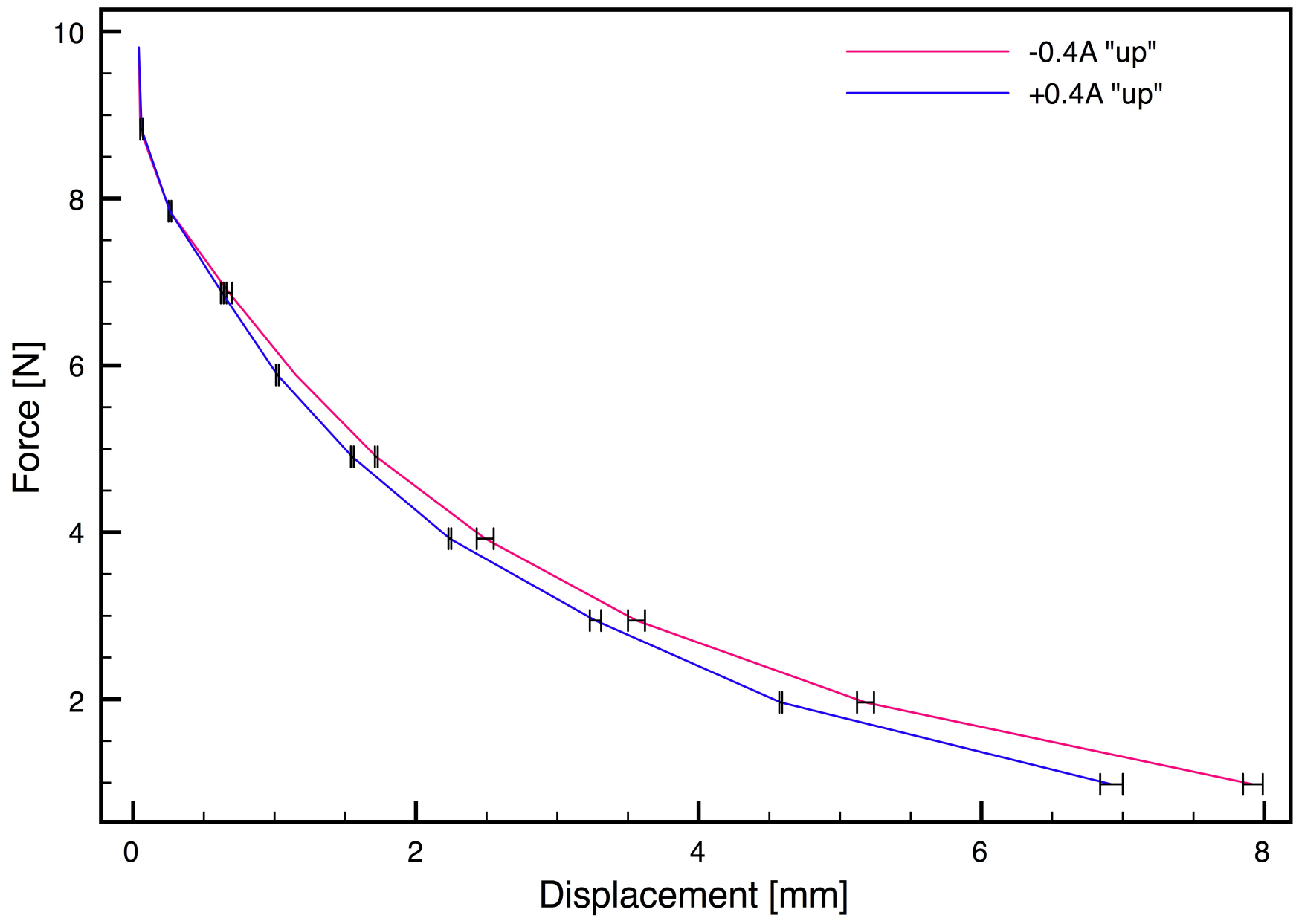

**Figure 2: Stiffness versus current.** Experimental results showing the change in the curve of the spring for two values of current (maximum ratings). The change in displacement is approximately 10%.

*Jack spring*

The principle in this tunable spring is long known. In Figure 3 (top), the basics are illustrated. A spring with a given stiffness is blocked half-way with a jack, such that only a part of the string deforms under the external force. The shorter this working part, the stiffer the spring (limited by the dimensions of the spring). The position of the jack is usually changed by screwing in or out. The standard version of this device (for an application see section Rumbo) has the major drawback that the length of the entire system changes when the stiffness is changed. To solve this problem we have designed the device shown in Figure 3 (bottom), originally proposed by Martin Kunz, an undergrad student. The idea is that the length that is screwed out in one side, is screwed in on the other, hence keeping the length constant (though mass distribution may change). This feature makes the device easy to mount in various situations, as in the WandaX swimming platform or in the quadruped Puppy. It also has the advantage that a motor can be mounted to produce the rotation that adjusts the stiffness of the spring. This actuation could be done on-line, provided that the angular velocity of motor is faster than the frequency of the external force. We are currently developing a first prototype and we expect to produce the working specification in short term.

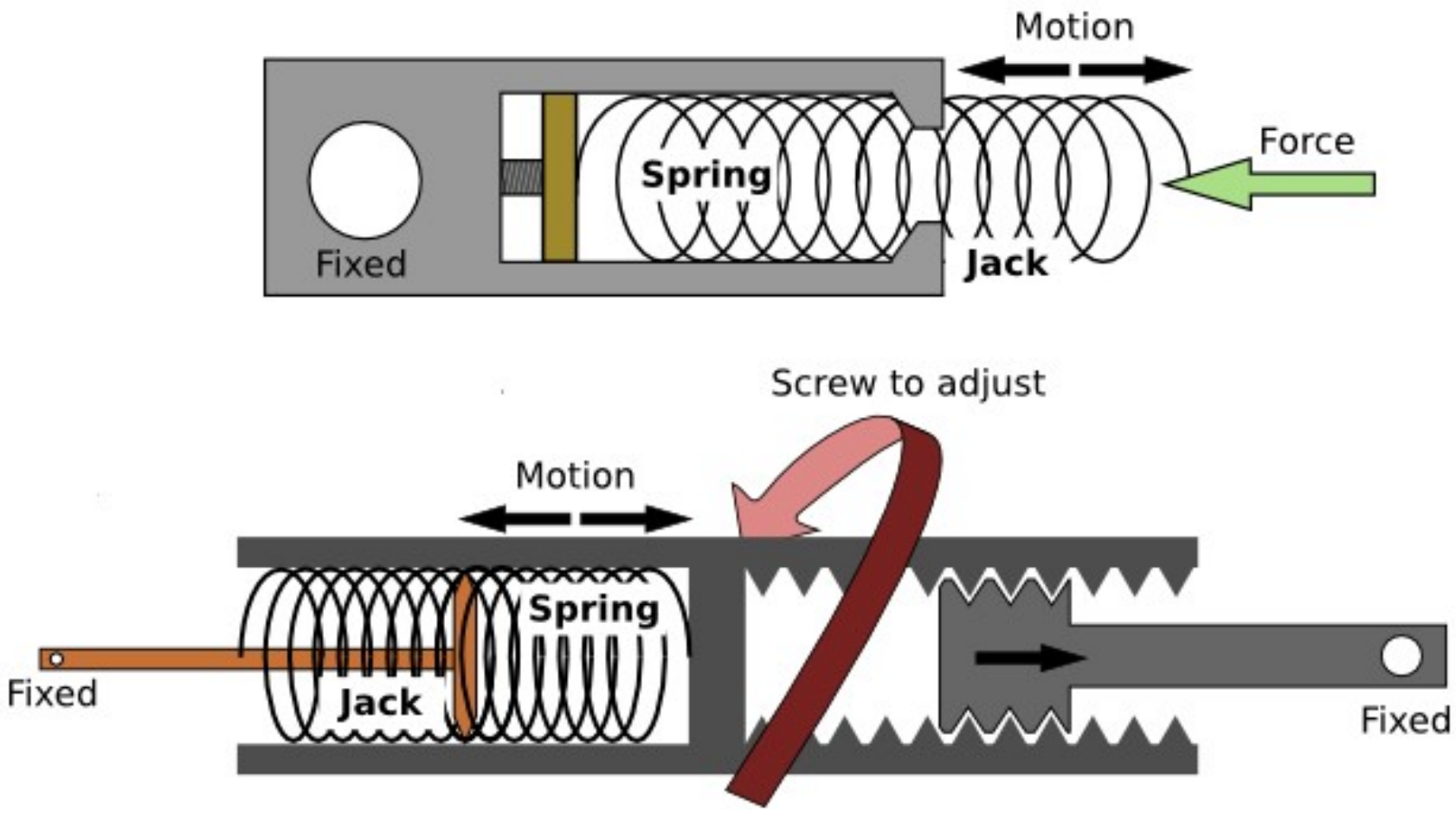

**Figure 3: Fix-length jack-spring**. This device is easier to mount than the usual design shown on top. It allows the installation of a motor to provide stiffness adjustment.

## Legged platforms

In this section we present results obtained in three platforms designed to study different aspects of legged locomotion. First, the "Zürihopper" platform served the purpose of investigating hopping behaviors with springs of different stiffness profile. We have studied how a tunable pneumatic spring could be used to keep the overall stiffness of a robot-ground system constant and at resonance. The second platform that has been used to study resonance phenomena was Rumbo. Third, we have continued our studies of morphology in quadrupedal robots and developed a new optimization framework for that purpose.

### *Zürihopper*

"Zürihopper" is an experimental platform designed and built by Emanuel Benker during his Bachelor's thesis work at our laboratory (Benker 2009). Zürihopper is a monoped (one-legged robot) and was designed to provide a tool to study hopping behaviors. During hopping, the platform behaves like a mass-spring system with flight phase, i.e. the system can completely hoist itself from the ground. The Zürihopper allows us to explore the effects of having legs with different spring curves (force as a function of compression of the leg) and to design spring control strategies to achieve a desired hopping behavior. In the following sections, we will describe the platform in detail and show how the spring curve can be set on the platform; finally we report the results from an experiment performed to reproduce a behavior observed in human hopping.

**Description of the platform.** In Figure 4, three views of the platform are shown. On the side view, at the end of a double boom, a pneumatic spring is connected. On top of the pneumatic spring, there is a mass connected to a DC motor. The motor moves the mass up and down (Scotch yoke mechanism), and provides the energy to the hopper.

The double boom guaranties that the motion of the hopper is almost vertical; the longer the boom the more vertical the motion. We used a 1 meter boom, which allows a maximal horizontal displacement of the foot of 1.3 millimeters. The moving mass can be observed in the front view and the way it is connected to the DC motor in the detailed view.

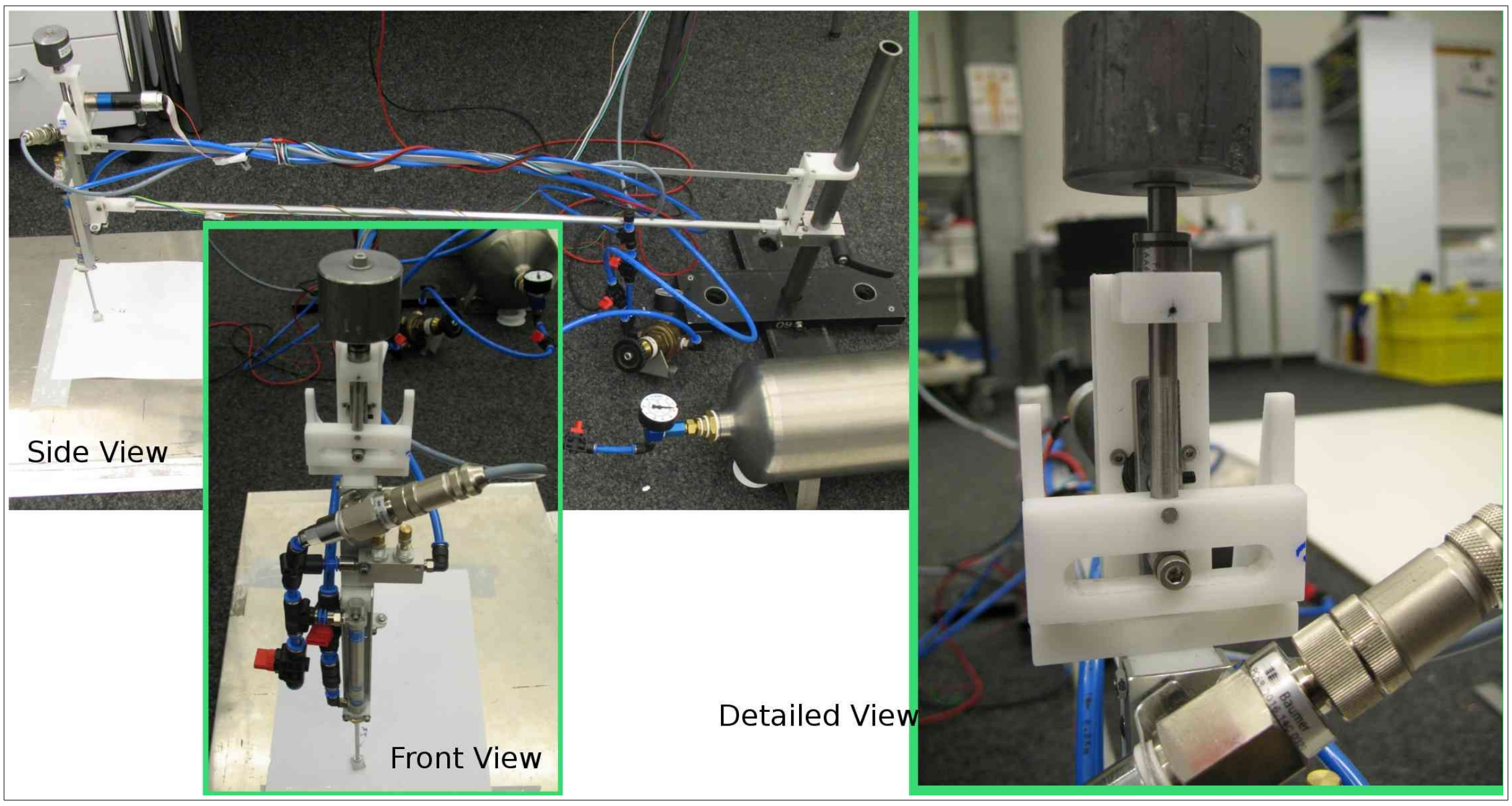

**Figure 4: Three views of Zürihopper.** The global view shows the complete setup. The way the mass is connected to the DC motor is observed in the detailed view.

The platform has two sensors: a pressure sensor inside the pneumatic spring and a distance sensor pointing downwards, towards the foot, as shown in Figure 5. The pressure on the spring was not controlled, but set to a given value at the beginning of each experiment. The angular velocity of the motor was controlled using a National Instrument DAQ board. The measurements from the pressure and the distance sensors were acquired with the same board.

**Setting the spring curve.** The spring curve can be estimated using the Ideal gas law. We assumed that the hopping process does not change the temperature of the gas, therefore

$$p_0 \cdot V_0 = p \cdot V$$

Using the notation of Figure 5, we obtain the force acting on the piston as a function of the compression:

$$F = \frac{\alpha}{\beta - A_v \cdot x} - \frac{\gamma}{A_v \cdot x + C_r}$$

$$\alpha = A_v \cdot P_{v,0} \cdot (V_{v,0} + C_v)$$
$$\beta = C_v + V_T$$
$$\gamma = A_r \cdot P_{r,0} \cdot (V_{r,0} + C_r)$$

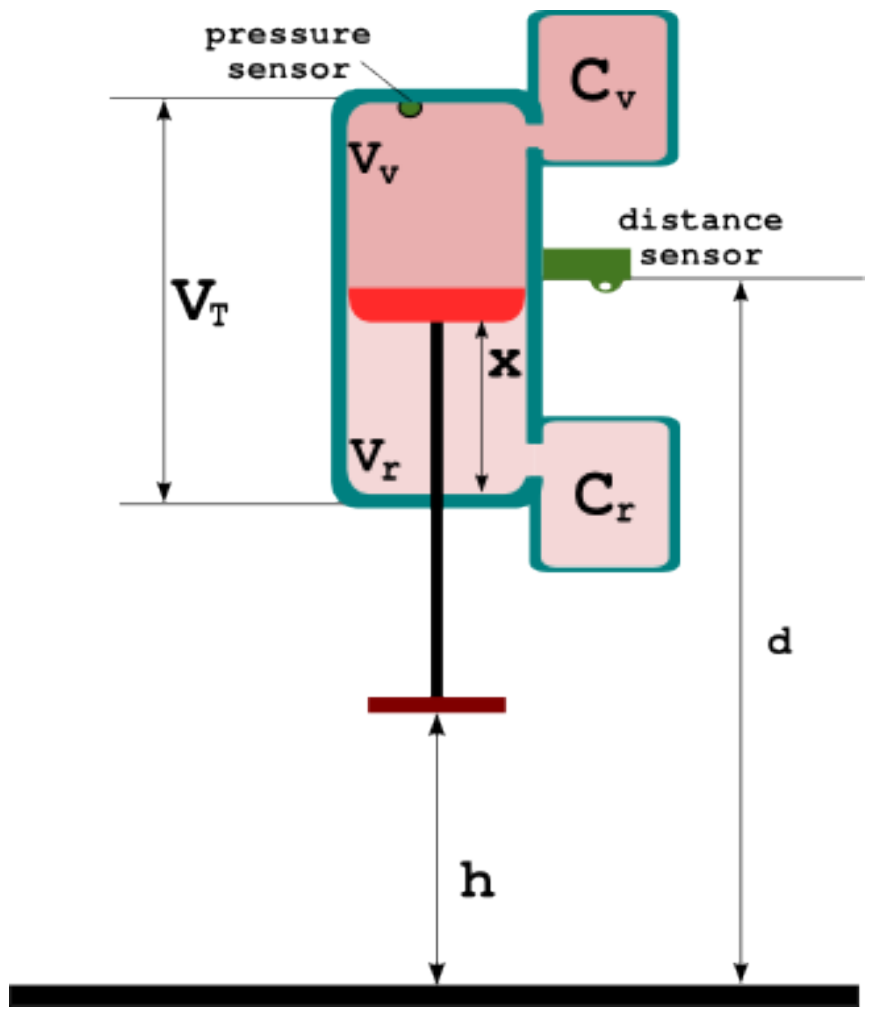

**Figure 5: Diagram of the setup.** Two chambers separated by the piston. Each chamber is connected to an additional chamber that does not changes volume during the motion. The position of the sensors is also shown in the diagram.

The letter A indicates the area of the piston facing the respective chamber. $P_{i,0}$ is the initial value of pressure, set when the system is in rest position. $V_T$ is the total volume of the cylinder. The constants $C_i$ are volumes that the piston cannot reduce while moving, we call them "dead volume". From the equation we see that the dead volumes could be controlled. By changing volume $C_v$, one could produce the spring curves shown in Figure 6. The advantage of controlling this parameter is that not only the slope of the spring curve is changed, but also the concavity. The values used to calculate the curves correspond to the current platform and are shown in the table below.

| Parameter | $A_v$ | $A_r$ | $P_{r,0}$ | $P_{v,0}$ | $V_{r,0}$ | $V_{v,0} = V_T$ | $C_r$ |
|---|---|---|---|---|---|---|---|
| Value | $7.9 \times 10^{-5}$ m$^2$ | $6.6 \times 10^{-5}$ m$^2$ | $3.0 \times 10^5$ Pa | $2.5 \times 10^5$ Pa | 0 m$^3$ | $6.3 \times 10^{-6}$ m$^3$ | $4.0 \times 10^{-6}$ m$^3$ |

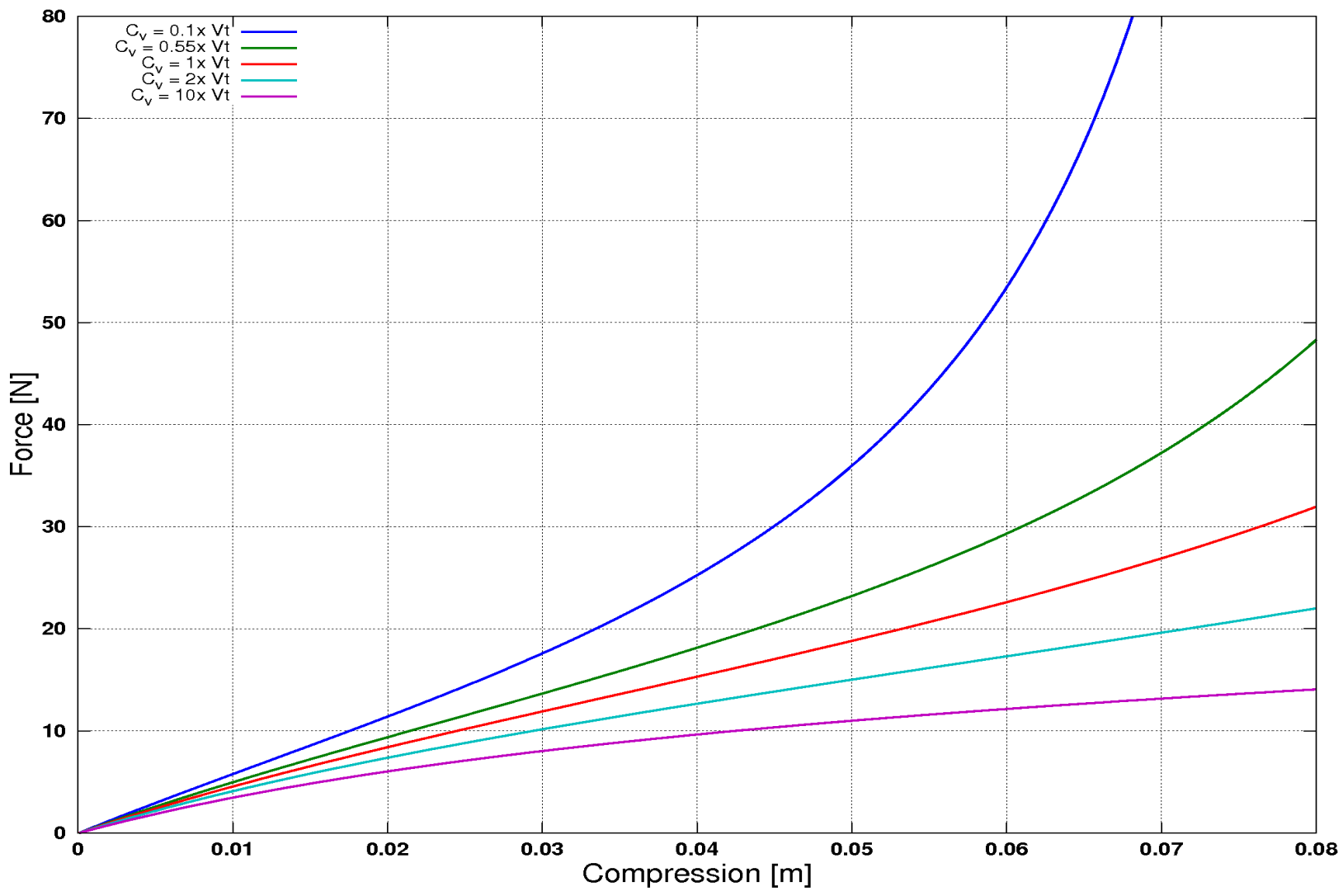

**Figure 6: Spring curves for different values of dead volume** connected to the upper chamber. This parameter changes the behavior of the spring from positive concavity to negative concavity.

Please note that we are not considering the effects of the flow of gas between the two cylinders and the dead volume chambers; the curves shown are equilibrium curves. The lower the maximum caudal that can flow between the cylinder and the chambers, the higher the damping on the system.

**Experiment: Keeping up with resonance.** In a paper from P. Ferris[2], it is observed that during hopping humans tend to adapt their leg elasticity in such a way that the total elasticity of the system body – ground remains constant. Using the Zürihopper we studied how the resonance frequency of the system changes with the ground elasticity. In this way, we are able to design a control strategy that produces the behavior observed by Ferris. The system used to control the stiffness of the ground is shown in a sequence of pictures in Figure 7. It exploits the bending of a beam with one extreme fixed and a sliding condition on the other. The longer the distance between the two supports, the lower the stiffness of the ground, where the Zürihopper lands.

We can estimate the amplitude of the jump by applying the Hilbert transform to the sensor signals. This gives a robust estimation of the amplitude even in situations where small beats are observed. We fixed the stiffness of the ground and calculated the amplitude of the oscillation for different frequencies. In this way, we were able to find the resonant frequency of the system, defined as the frequency where the amplitude is maximum. The results are shown in Figure 8. Repeating the same process with different pressures on the upper chamber ($P_{v,0}$), we observed a hysteresis-like phenomenon; the frequency at which the Zürihopper starts jumping depends on the direction of change of the frequency. Also, a double peak is observed due to the fact that multiples of the resonant frequency can also induce jumping. The higher the pressure, the higher the resonant frequency, as expected. Interestingly, the width of the hysteresis-like loop is increased with higher pressures.

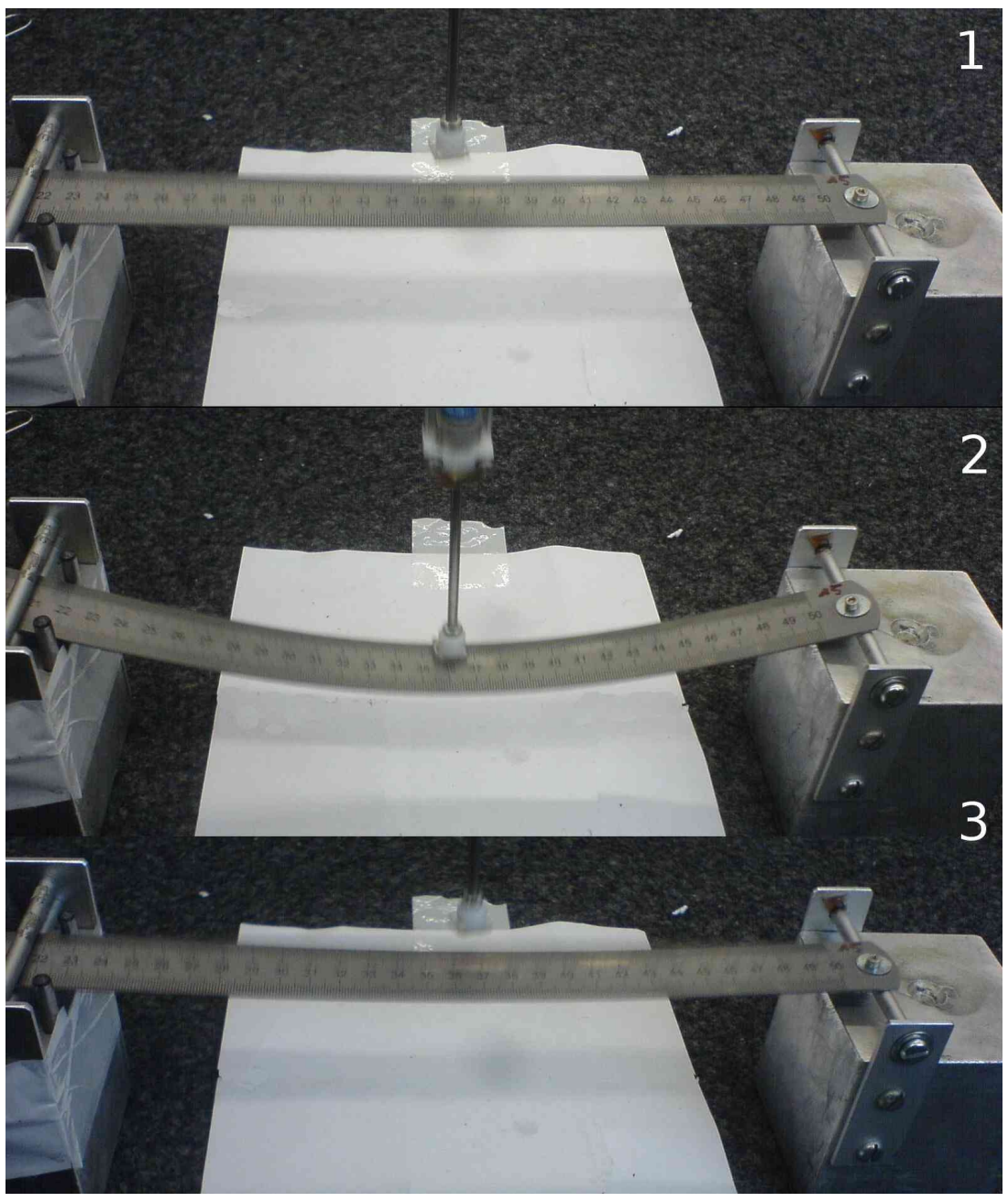

**Figure 7: Ground with variable stiffness.** This setup exploits the bending of a metal beam to produce stiffness depending on the length of the beam.

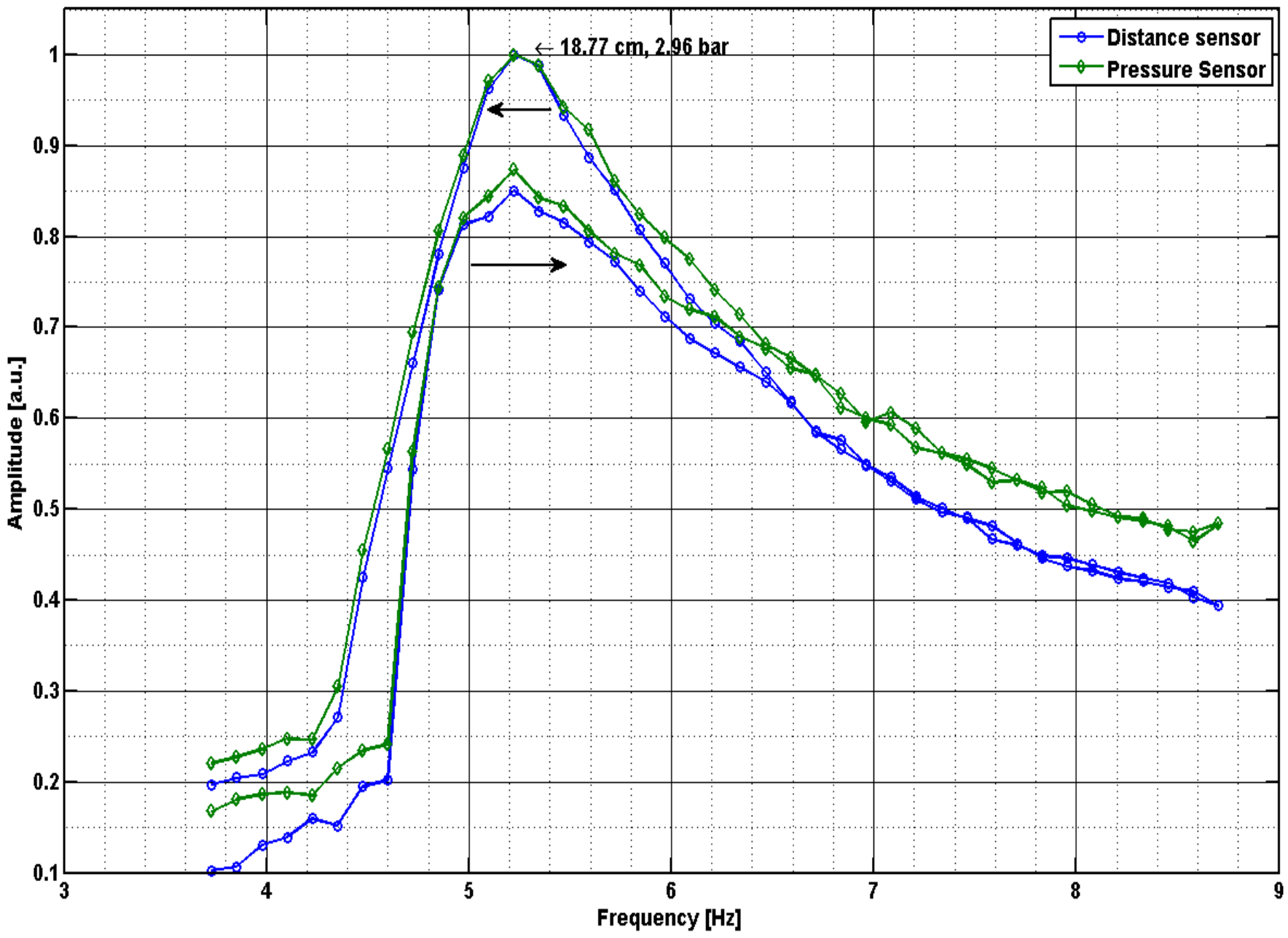

**Figure 8**: **Amplitude versus frequency.** There is a frequency for which the system shows maximum amplitude. Depending on how the frequency is varied, the value at which this maximum happens can change. The arrows indicate increasing and decreasing frequency.

[2]Ferris, D. & Farley, C. (1997), Interaction of leg stiffness and surface stiffness during human hopping, Journal of Applied Physiology 82, 15-22.

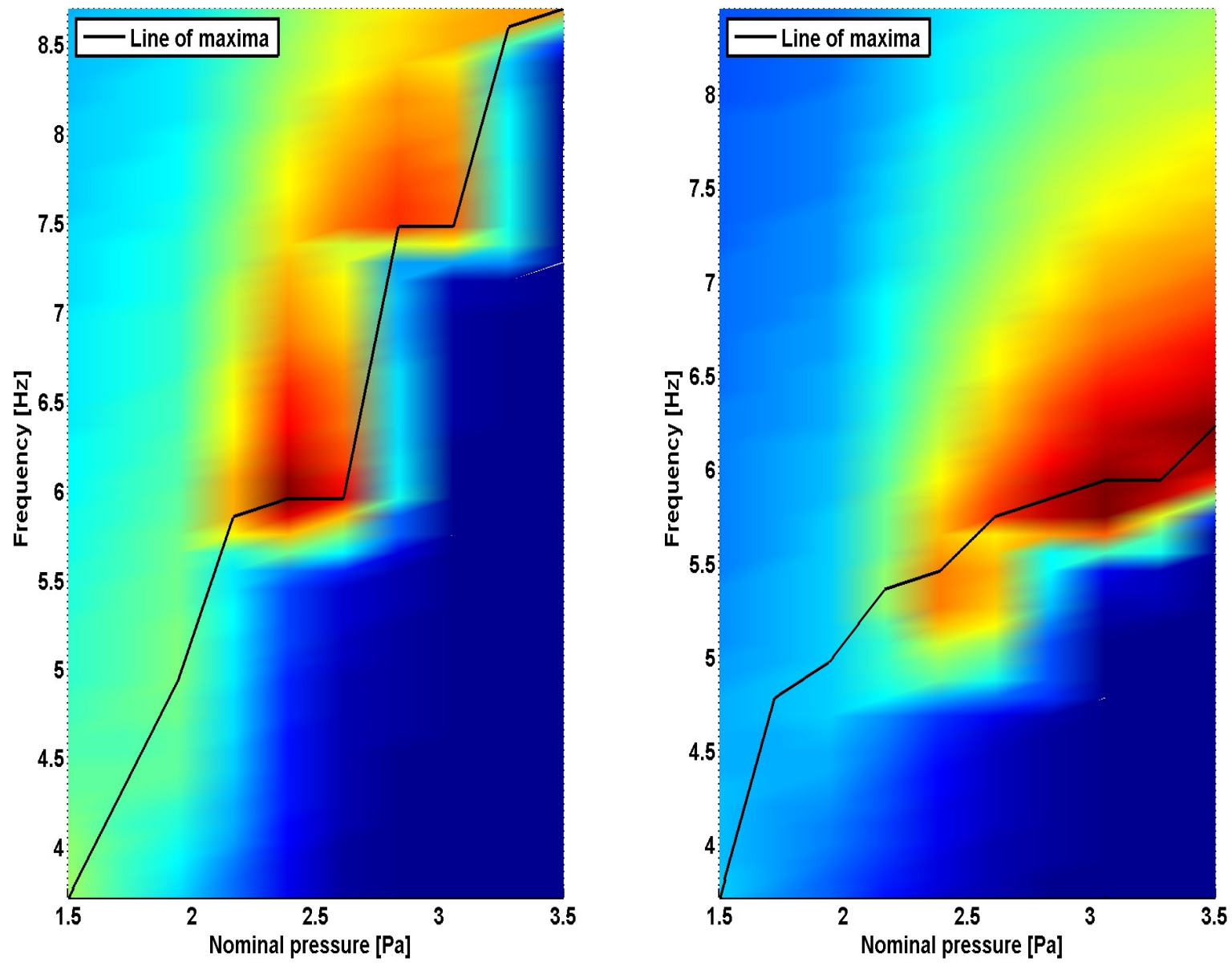

**Figure 9: Amplitude in the frequency-pressure plane**. The line of maxima is marked in black. A controller should follow such a trajectory while keeping the system in resonance.

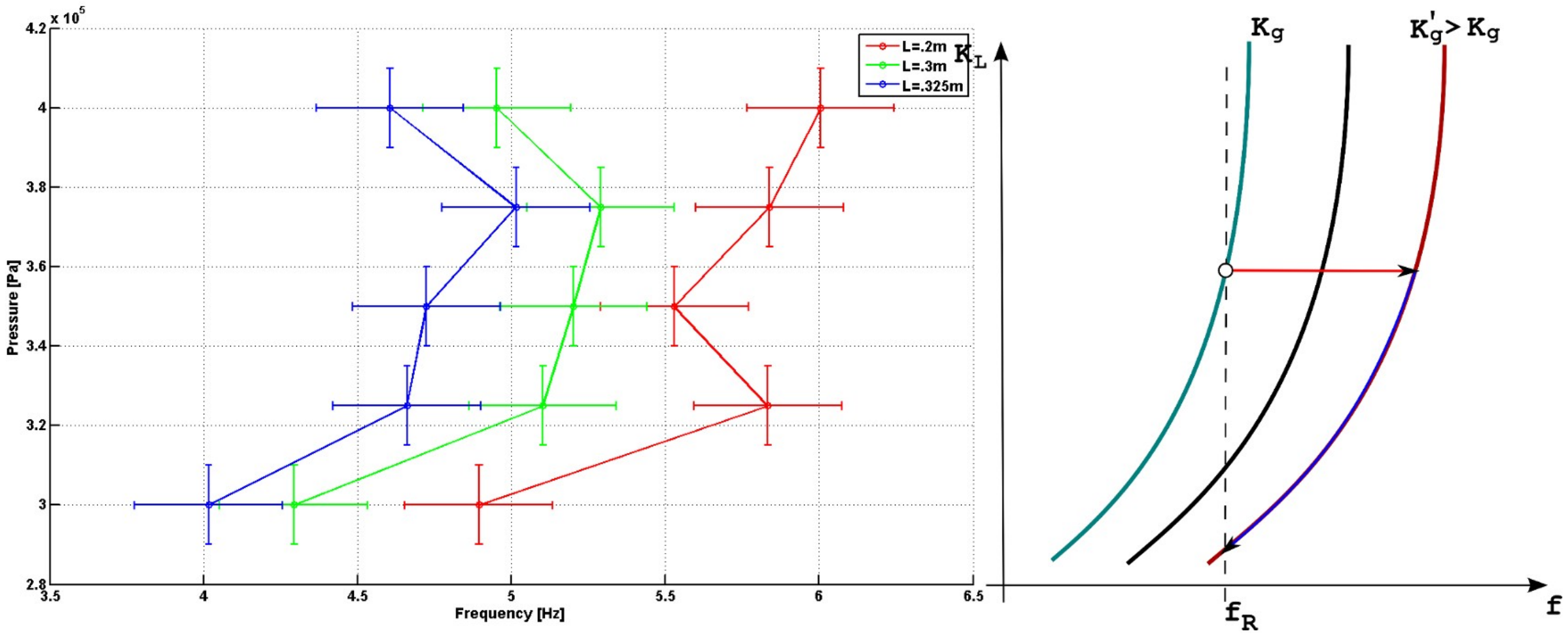

**Figure 10: Change of resonance frequency as a function of ground stiffness.** The distance between the supports increases from right to left. The longer the distance, the lower the resonant frequency. On the right a possible adaptation strategy is shown**.** The ground increases its stiffness ($K_g$) and the robot does not control the frequency of the mass ($f_R$). In this situation, the robot would observe a decrease in its amplitude due to the change of the resonance frequency of the robot-ground system (red arrow). Following the change of the amplitude, the leg's spring stiffness is reduced such that the system is back into resonance (blue arrow).

The amplitude value in the frequency-pressure plane is shown in Figure 9. We have marked the line of maximum values; this line represents the ideal trajectory of a pressure controller that is trying to adapt to changes in the robot-ground system, in our case ground stiffness. The effect of the ground stiffness on the resonance frequency was measured, the results are shown in Figure 10 (left). On the right of the figure, the behavior of such a controller is depicted. Let us take for illustration purposes the case when the floor increases its stiffness and the robot does not control the frequency of the mass. In this situation, the robot would observe a decrease in its amplitude due to the change of the resonance frequency of the robot-ground system. However, following the change of the amplitude, the leg spring stiffness is reduced such that the system is back into resonance.

### *Rumbo*

One of the most popular robots in our laboratory during "Lab Tours" is Dumbo. It is made of a flexible body and two unbalanced motors. When the motors are turned on, the structure goes slowly into resonance and starts moving forward, resembling a primitive way of walking. One of the questions that were open was whether it is necessary to have a flexible body to show this behavior or if a rigid structure with one springy joint could also do it. To answer this question, we built Rumbo, a rigid version of Dumbo. The platform is shown in Figure 11 (top center). The upper part of the body is connected to the lower part (the foot) trough a  ball bearing joint and a tunable jack spring. The platform allows the adjustment of the length of the foot, the position of the center of mass, and the position where the jack spring is connected. The jack spring itself allows to set the stiffness of the joint (but not the rest position of the system). The platform indeed shows the walking like behavior of Dumbo. Additionally, by moving the center of mass horizontally, we found that the direction of motion is dependent on it. If the center of mass is at the left of the middle of the foot, the platform moves towards the right and vice versa. The results where verified by 2D physical simulations.

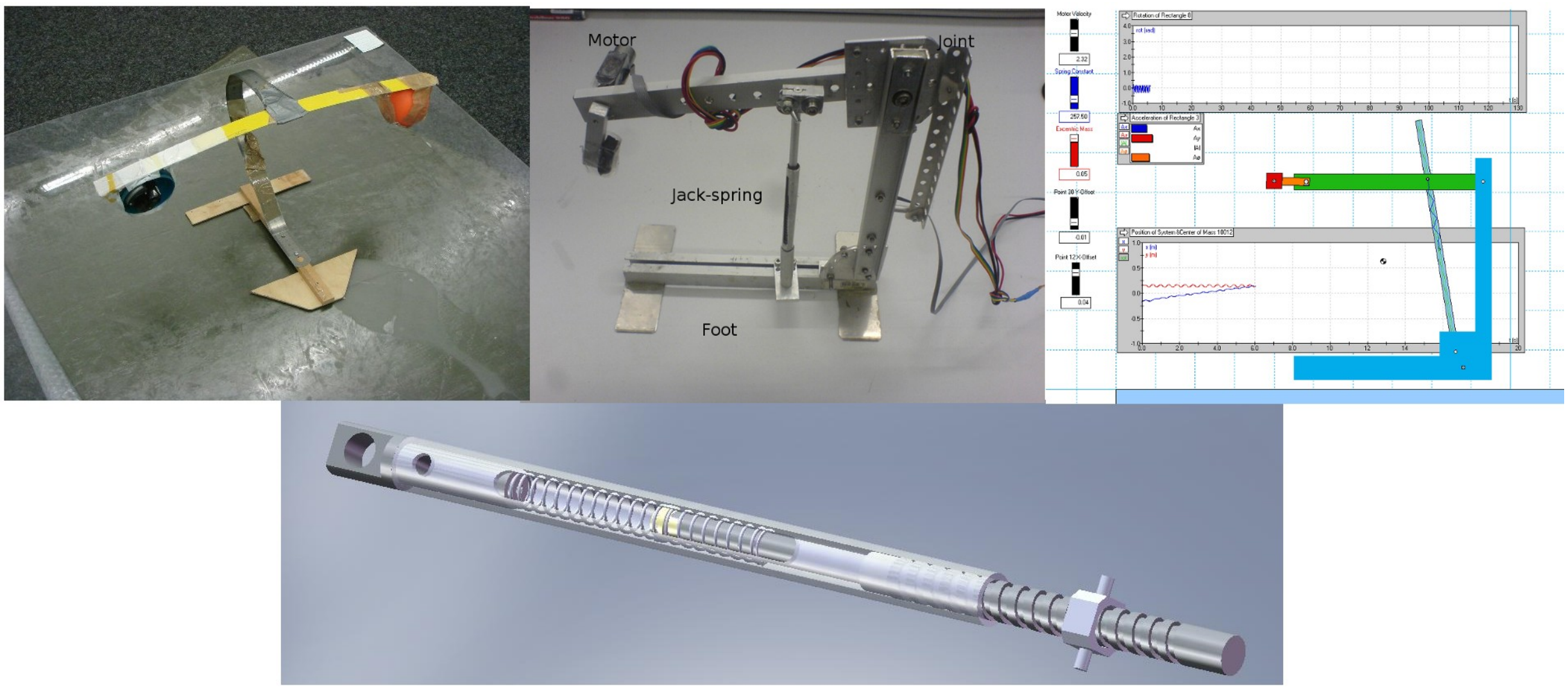

**Figure 11: Dumbo and its rigid counterpart**. On the top left a picture of Dumbo, to the right, Rumbo, the rigid version and a 2D model of it. On the bottom we show the CAD of the jack-spring.

### *Quadrupedal robots*

For our quadrupedal platforms, we have further developed the model of the robot in the Webots simulator[3]. First, in two BSc. theses (Hutter 2009, Faessler and Ruegg 2009), we have been co-evolving the morphology and control of quadrupedal robots engaged in different gaits. The following morphological parameters were varied: femur and tibia lengths for all legs, spring and damping constants for passive knee joints, center angles for passive knee joints (where spring is in rest position), and body length, width and center of mass. While at first we have searched for the fastest gaits, later we have introduced energy consumption to the fitness function and were thus able to search for the most energy efficient combinations of morphology and controller for different gaits: walk, trot and bound. We have arrived at distinct parameter combinations for different gaits (Fig. 12). However, the results require further analysis and verification on the real robot.

[3] http://www.cyberbotics.com/

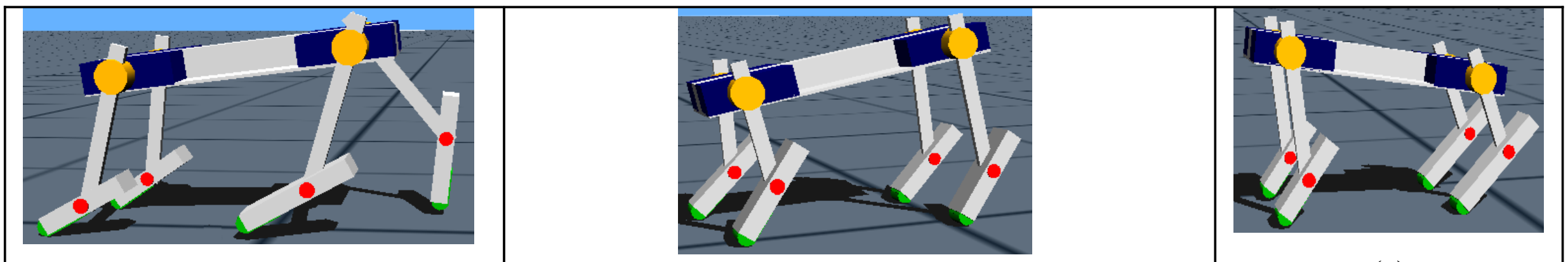

(a) (b) (c)

**Figure 12: Results of morphology optimization in Webots simulator.** Fittest individuals for walking (a), trotting (b), and bounding gait (c).

While we wanted to see, how specialized morphologies look like, we are also seeking 'compromise' morphologies that can accommodate diverse gaits. This and other requirements (we also wanted the gaits to be stable against control parameter variations in a certain range and we wanted to be able to change the speed of a gait by varying a single parameter - frequency) have led us to the development of a new modular optimization framework architecture Vidyya.[4] This allows to use any optimization algorithm and to perform even hierarchical optimization experiments. The last point we are currently addressing in the quadrupedal platforms is the issue of active vs. passive joints and joints that can switch between the two. We are starting experiments in the Webots simulator where individual joint characteristics are subject to optimization. A joint can be either passive (and then also its compliance is optimized), active (with a motor), or active/passive – switching between the modes on command (For instance, active control is applied only during leg lift-off to provide ground clearance. At other times, the motor is clutched and a passive compliant joint is used.) The optimization criteria are energy efficiency as well as maneuverability and performance on difficult terrain.

## Swimming platforms

We have focused on two swimming platforms: WandaX and Wanda2.0. WandaX continues our investigation into how thrust is generated with an undulating body. This platform allows us to explore the stiffness distribution along a passively compliant multi-segmented body. At the same time, the platform can be modeled in a fluid dynamics simulator. Wanda2.0, on the other hand, is a complete, autonomous robot that will be able to swim in open water. It is being equipped with a broad range of sensors and will feature a tail fin with on-line tunable stiffness.

### *WandaX*

The WandaX experimental robot platform was built considering two aspects: First, the obtained results have to be easily portable to a CFD (computational fluid dynamics) simulation. Second, the platform has to be easily reconfigurable to allow for various explorations of the effects of the robot morphology. Following the former aim, to reduce the complexity of the structure to be simulated and to save computation time, the robot body is simplified to a plate and the tail fin consists of four smaller rigid plates. Furthermore, the supporting platform allows the robot to swim only on the surface, which reduces the complexity to a 2D model. The CFD simulation allows us to evolve interesting morphologies much faster, test control strategies for actuation or evaluate the underlying mathematical model. The second aspect aspect involves the following: the position of the actuated axis on the robot, the tension of the passive

[4] http://vidyaa.origo.ethz.ch/

joints, and size of the plates in the water are reconfigurable.

**Description.** On top of the platform segments, a passive connector or an actuator can be mounted. At present, only one joint is actuated, and the remaining three are passive. This means WandaX is an under-actuated system. In Figure 13 a detailed view of the platform is shown and the notation used in this paragraph is defined. The robot consists of five segments connected through pairs of ball bearings. Each segment can hold a plate that is submerged into water. These plates can be changed in terms of size or material. In the current setup, the front plate (the actuated joint is in the middle of the robot) has the same surface as the total surface of all the four rear plates. The only asymmetry, which is crucial for the forward propulsion in this simplified setup, lies in the flexible connection of the passive joints in the tail fin. The modularity allows us to change the shape for further experiments in the following way: The plates submerged in the water can be replaced by ones of different size, changing significantly the forces between the surrounding water and the robot body.

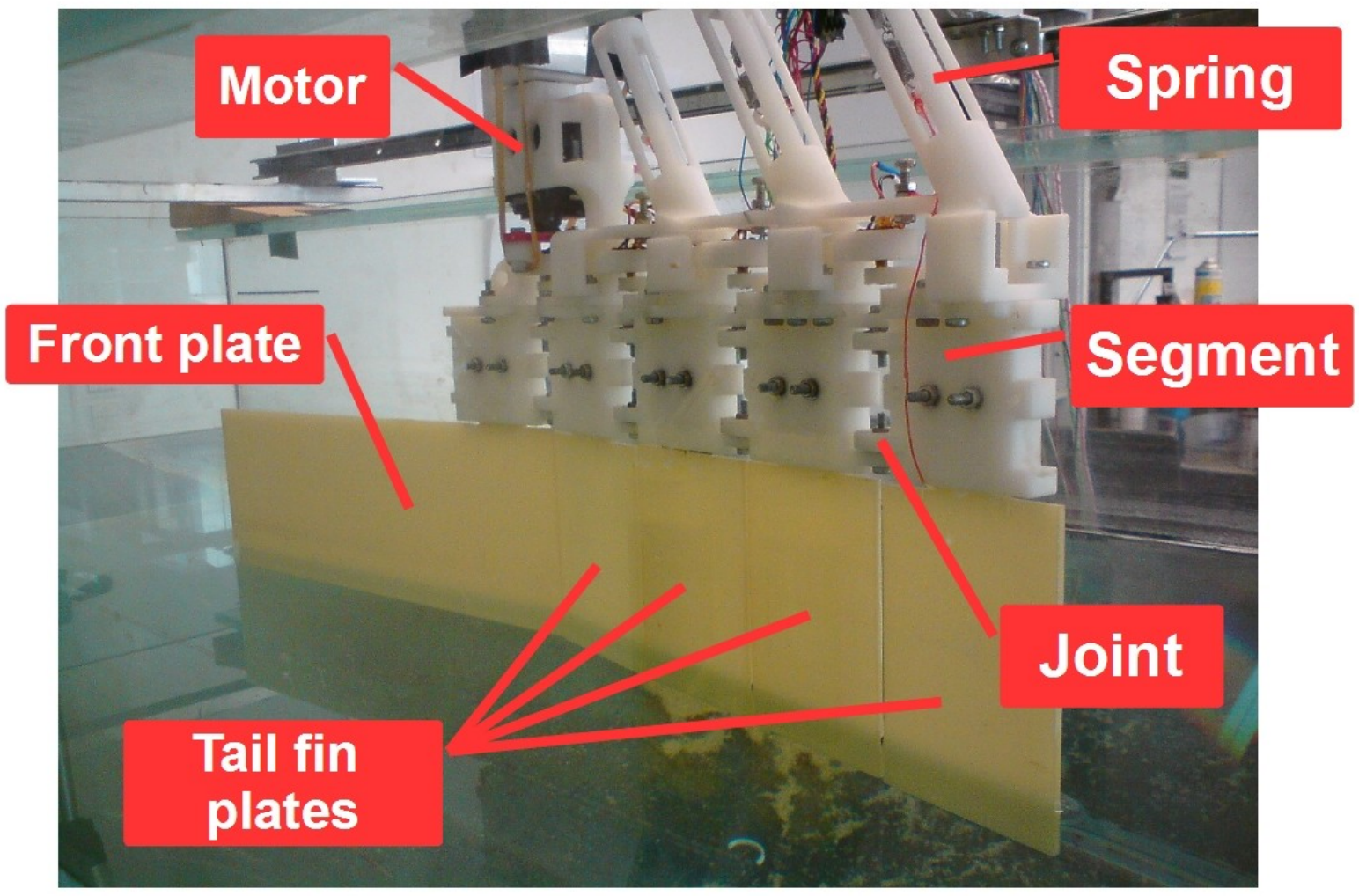

**Figure 13: The WandaX experimental robot platform** hanging from the frame and submerged around 4cm into the water. The most important parts are labeled with the terms used throughout the text.

The other three passive joints have linear springs that will stretch when the segments are moved out of their alignment, as shown in Figure 14. One side of the spring is connected to the spring tunnel fixed to one of the segments; the other side is attached through a string to a lever fixed to the neighboring segment. At rest position, the spring holds the segments aligned and when a force is applied to one of the segments, it deflects against the spring force. The design allows to change the stiffness of the individual passive axes. The position of all four joints (three passive and one actuated) are monitored as well as the power consumption of the motor and the generated thrust in one direction (x axis). Experiments are conducted with different parameters (frequency and amplitude) to the servo motor (Ziegler and Carbajal, in preparation).

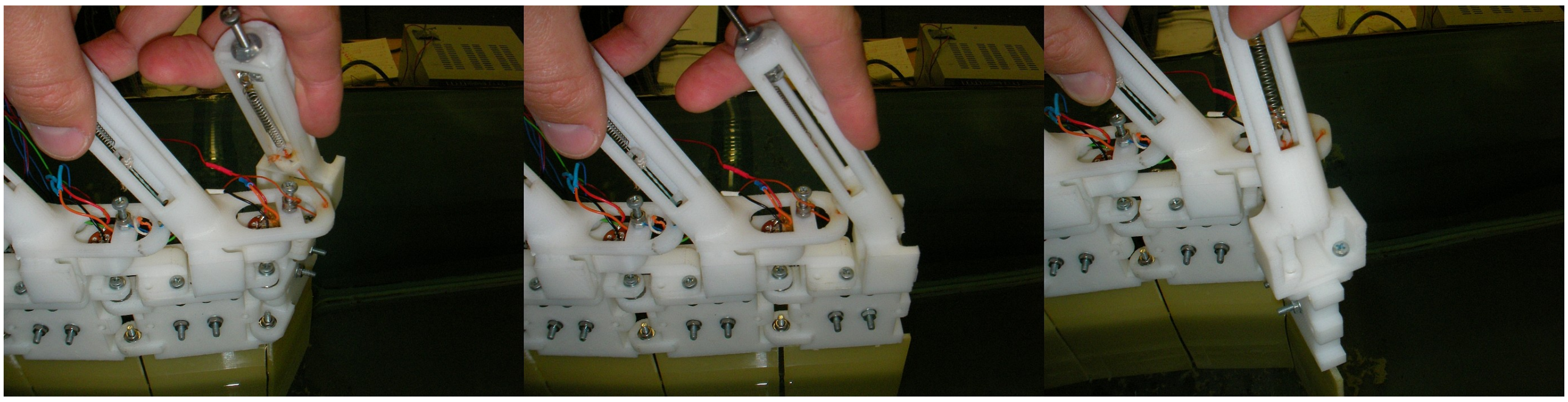

**Figure 14: Tunable spring in WandaX**. From left to right: the last segment is manually bent from the leftmost to center and rightmost position. The spring is extended more whenever the segment is not aligned with its neighboring one. The screw on top of the “spring tunnel” can be screwed in to shorten the spring. The force holding the segments aligned is then reduced and less force is needed for the same deflection.

Currently, the pretension of the springs is changed manually using a screw holding one side of the spring (off-line tunable spring). The screw that holds the spring can be replaced by a little motor and a lever attached to it, providing on-line tuning. Depending on the lever position, the spring has then more or less pretension and consequently stiffens the passive joint. This allows the robot to change its morphological properties (here the stiffness of the tail fin segments) while running. This new ability of the robot raises a number of questions which we try to answer: What is the best stiffness distribution along the body for generating maximum thrust or highest efficiency? Is there only one optimal distribution or is it depending on the current behavior and environment? Is it beneficial to change stiffness within one stroke of the tail fin? What is a good control architecture that can take the ability of changing morphological properties into account?

### *Wanda2.0*

Wanda2.0 is a new swimming platform that we are currently developing. With previous Wanda robots, we have shown that a fish robot with a single degree of freedom can swim in 3 dimensions and that the material properties of the tail fin play a critical role. In the new platform, we want to build on these insights and extend them in the following manner: (i) We are adding a tail fin whose stiffness can be manipulated on-line; (ii) Wanda2.0 is equipped with many sensors encompassing multiple modalities. This will allow the robot to acquire more information about its interaction with the environment. We will finally be able to test more advanced control algorithms (that include on-line manipulation of morphology), and work on exploration, and exploitation of morphology, body schema synthesis and forward modeling (WP 2 and 3) also on a swimming platform.

Since Wanda2.0 is a free swimming robot platform, special care on designing and selecting the components was taken. First, Wanda2.0 is an untethered platform with an on-board micro controller. However, for control as well as analysis of its behavior, we are implementing a wireless communication with an external computer. Second, for the architecture of the on-board electronics, we chose a decentralized architecture. Three micro controllers are reading and processing the sensors and running the servo motors. This provides higher performance, redundancy and future extensibility. Currently, all the electronic components, including batteries and motors, are wired and tested before placing them into the final robot.

Wanda2.0 has a three axis accelerometer, two one axis gyroscopes, power monitor, bending sensors in the tail fin, compass module and a water pressure sensor. The information of orientation and depth over time can be used as a measure of fitness while searching in the parameter space frequency – amplitude – offset – elasticity of the controller for good combinations. An important aspect is to keep the possibility to freely place most of the sensors anywhere on the body of the robot, since we are planning to explore how the orientation of the sensors is correlated to the kinematic information and consequently how they influence the different control architectures.

**Fin with tunable stiffness.** The stiffness of the tail fin is one of the crucial factors that influence speed, energy-efficiency and behavioral diversity of a swimming platform. Therefore, it is of particular interest to vary the stiffness on-line. Since we cannot change Young's module of the material used in the fin, the stiffness will be varied by inserting foil stripes from different materials into the fin structure. Wanda2.0 has two servomotors for positioning the stripes in the tail fin. Depending on how deep each of the stripes is inserted, the overall stiffness will change. Furthermore, the flexibility is not necessarily homogeneous along the tail fin but can vary from stiff (close to actuator) to compliant (tip of the fin).

## (b) Work package 2: Learning to exploit body dynamics

The goal of this work package is that a robot can automatically explore, learn about, and then exploit the action possibilities it has, given a particular body and environment. These components are very tightly intermingled. Nevertheless, we will structure this part of the report as follows: First, a section on exploration and exploitation of body dynamics will on one hand address various control architectures that can facilitate this goal, and on the other hand it will address how the 'landscape of possibilities' can be analyzed through modeling. Second, we will report some results on how can a robot structure and represent the information it has acquired during the exploration phase – leading to a synthesis of a primitive body schema.

### Exploration of body capabilities and learning to exploit them

The goal we have set ourselves for a locomotion controller is that it has to discover the capabilities of a particular robot body interacting with its environment and learn to exploit rather than override the 'natural modes' of interaction. In a series of Master and Bachelor theses, we were exploring different solutions to this problem in quadrupedal locomotion. In Hutter (2009), and Faessler and Ruegg (2009), we have explored the limits of a feed-forward controller when co-optimized together with the robot morphology. In Nuesch (2009) and Michel (2009), we have investigated two different control architectures where oscillators are coupled to the body through feedback connections and, under certain circumstances, get entrained to the resonant frequencies of the body-environment system. As yet another approach, we are currently investigating a controller featuring chaotic dynamics. At last, to learn about the possible normal modes that are to be discovered by the controllers, we have engaged into modeling our quadrupedal platforms.

#### *Co-optimizing a feed-forward controller and robot morphology*

In Hutter (2009), and Faessler and Ruegg (2009), we have explored the limits of a feed-forward controller when co-optimized together with the robot morphology (see also Quadrupedal robots section in WP1). With this approach, a controller capable of driving the robot to stable and energy-efficient locomotion can be found. However, we see two drawbacks: (i) the exploration process is long; and (ii) the controller cannot adapt to changing circumstances. Therefore, we have devoted more effort to exploring controllers that incorporate feedback.

#### *Entrainment with feedback oscillators*

Feedback controllers can provide a remedy to both problems. Feedback can be used to channel the exploration process as well as to make a controller adaptive.

**Adaptive Hopf oscillator.** In Nuesch (2009) we have investigated one particular instance of a feedback controller with the above-mentioned capabilities: the adaptive frequency Hopf oscillator. This oscillator is used to drive the actuators, but through feedback connections, it is able to tune itself to resonant frequencies in a mechanical system. We wanted to gain further insights into the work done previously on the Puppy II platform[5] (which can be seen in Fig. 19 or Fig. 20). First, however, we wanted to clarify the behavior of the oscillator in a closed loop under controlled conditions. For that we have built a simple simulator: a harmonic mass-spring oscillator is in place of the mechanical system and is connected to the Hopf oscillator (Fig. 15 a)). There are several important parameters that affect: (i) whether the Hopf

[5]Buchli, J.; Iida, F. & Ijspeert, A. J. (2006), Finding resonance: Adaptive frequency oscillators for dynamic legged locomotion, *in* 'Proc. of the IEEE/RSJ International Conference on Intelligent Robots and Systems (IROS 06)', pp. 3903-3909.

Buchli, J. & Ijspeert, A. J. (2008), 'Self-organized adaptive legged locomotion in a compliant quadruped robot', *Autonomous Robots* **25**, 331-347.

oscillator's frequency converges; (ii) the value to which this frequency converges (note that this is not always the resonant frequency of the mechanical system); (iii) the speed of convergence. The parameters that we investigated in Nuesch (2009) are: (i) initial difference between frequency of Hopf and harmonic oscillator; (ii) feedback gain; (iii) phase shift on the feedback connection. All of these parameters have significant effects; interestingly, the phase shift introduced to the feedback connection has a critical impact (Fig. 15 b)).

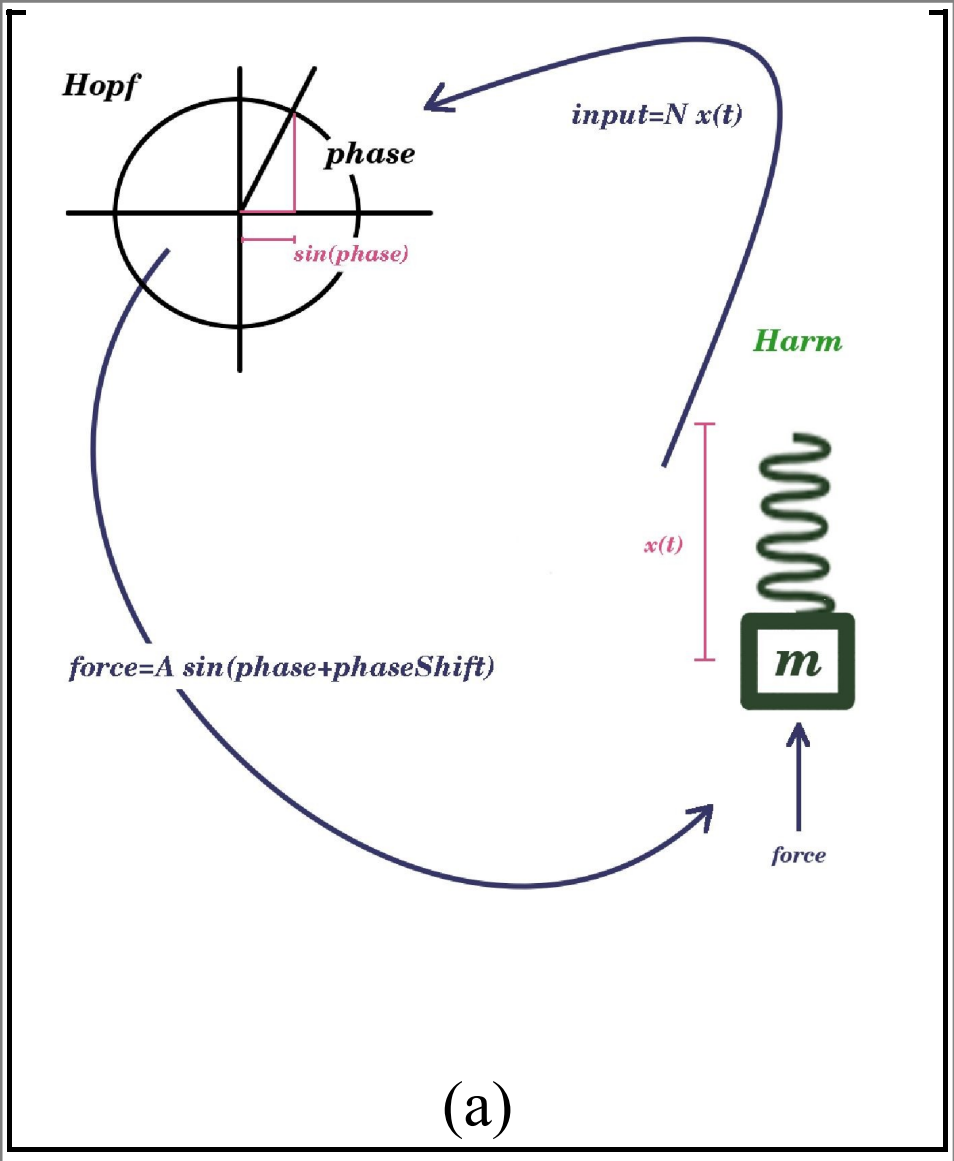

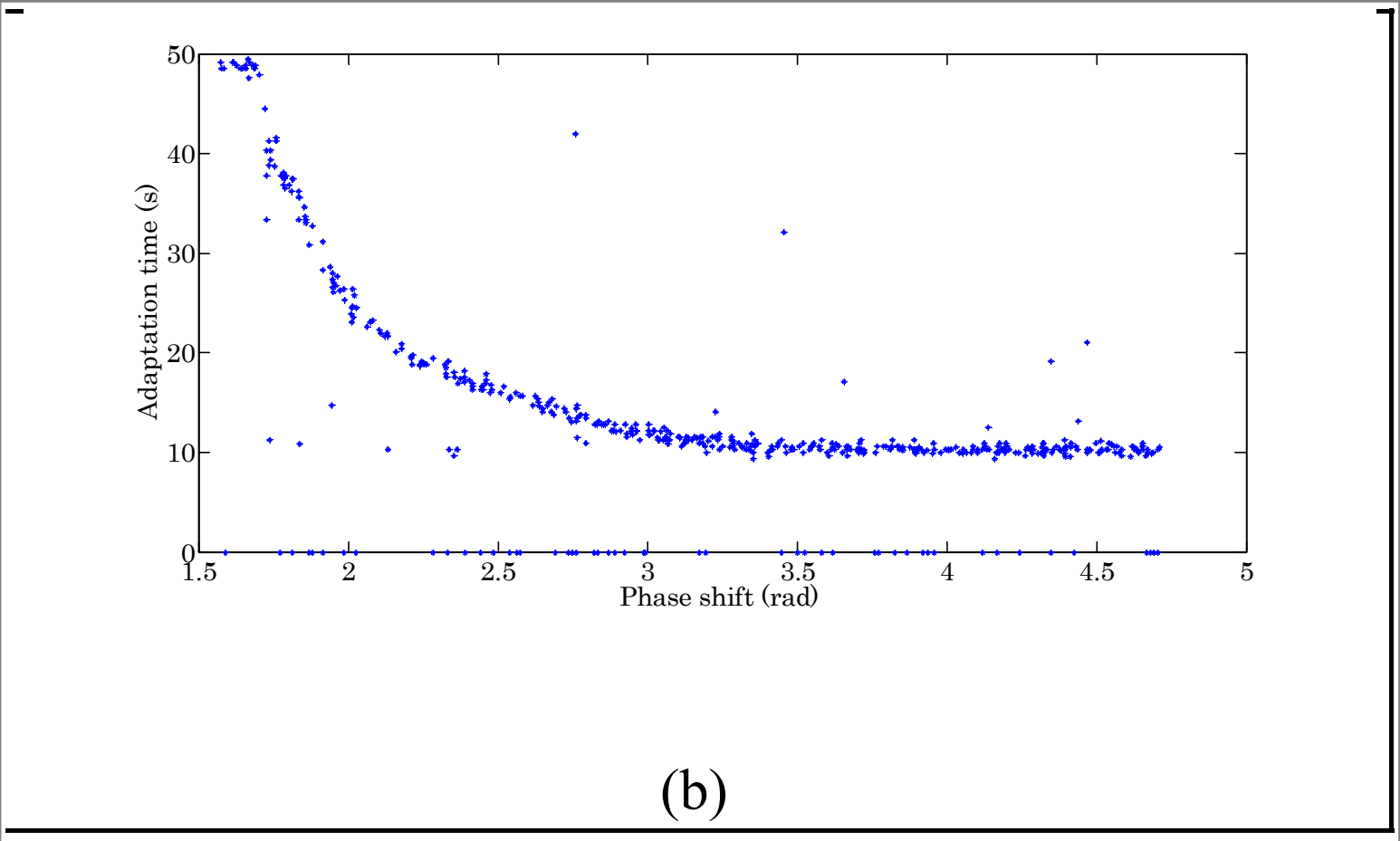

**Figure 15. Adaptive frequency Hopf oscillator coupled to a harmonic oscillator.** (a) Schematics of the setup. The Hopf oscillator acts as a controller and is sending a forcing signal to a simulated mass-spring harmonic oscillator. Its amplitude is then sent as input to the Hopf. (b) An experiment with phase lag introduced onto the feedback signal to the Hopf oscillator. This significantly affects the time of frequency adaptation.

We have then transferred our controller (the adaptive Hopf oscillator) to the real quadrupedal robot. The Puppy II robot has 4 passive knee joints with springs. If these get excited close to their natural frequencies, the energy transfer from the motors is more efficient. However, the robot as a whole is a more complex system, with multiple resonant frequencies. These in turn also depend on the posture and gait, for instance. The first problem to attack was all four legs oscillating synchronously. In previous work on the Puppy II robot[5], convergence of frequency adaptation was achieved with vertical acceleration sensor as feedback, but not with angular sensors in the knees of the robot. In Nuesch (2009), we have achieved convergence with both sensors. However, the frequency the oscillator has converged to was substantially different for the two sensors and also dependent on the phase lag that we have been varying. We have tried to explain these phenomena in follow-up work (see also section on Modeling normal modes), where we have introduced a more detailed model of the robot and its sensors. We have considered the effect of damping, centre of mass position, as well as of a flight phase (see also Zürihopper). Still, many open questions remain.

**Synchronization in oscillator communities.** As another controller consisting of oscillators that are coupled to feedback, in Michel (2009) we have studied oscillator communities (a set of interconnected dynamical systems with periodic behavior) of Kuramoto oscillators. We used the following setup in simulation. A robot has four legs with elastic passive knees (rotational springs). Each leg of the robot is actuated at the hip by a single servo motor. Each motor is connected to an oscillator community. The four communities are not explicitly connected to each other. However, each community receives local inputs from the angle sensors situated at the knees of the robot (see Figure 16). It is expected that for some configurations of the setup, synchronization among the four communities will occur and, as a consequence, emerging gaits will be observed. This contrasts with the standard approach where a gait is pre-designed and realized in a network of strongly coupled oscillators whose pattern is dictated to the body. Nevertheless, we have not obtained any conclusive results yet.

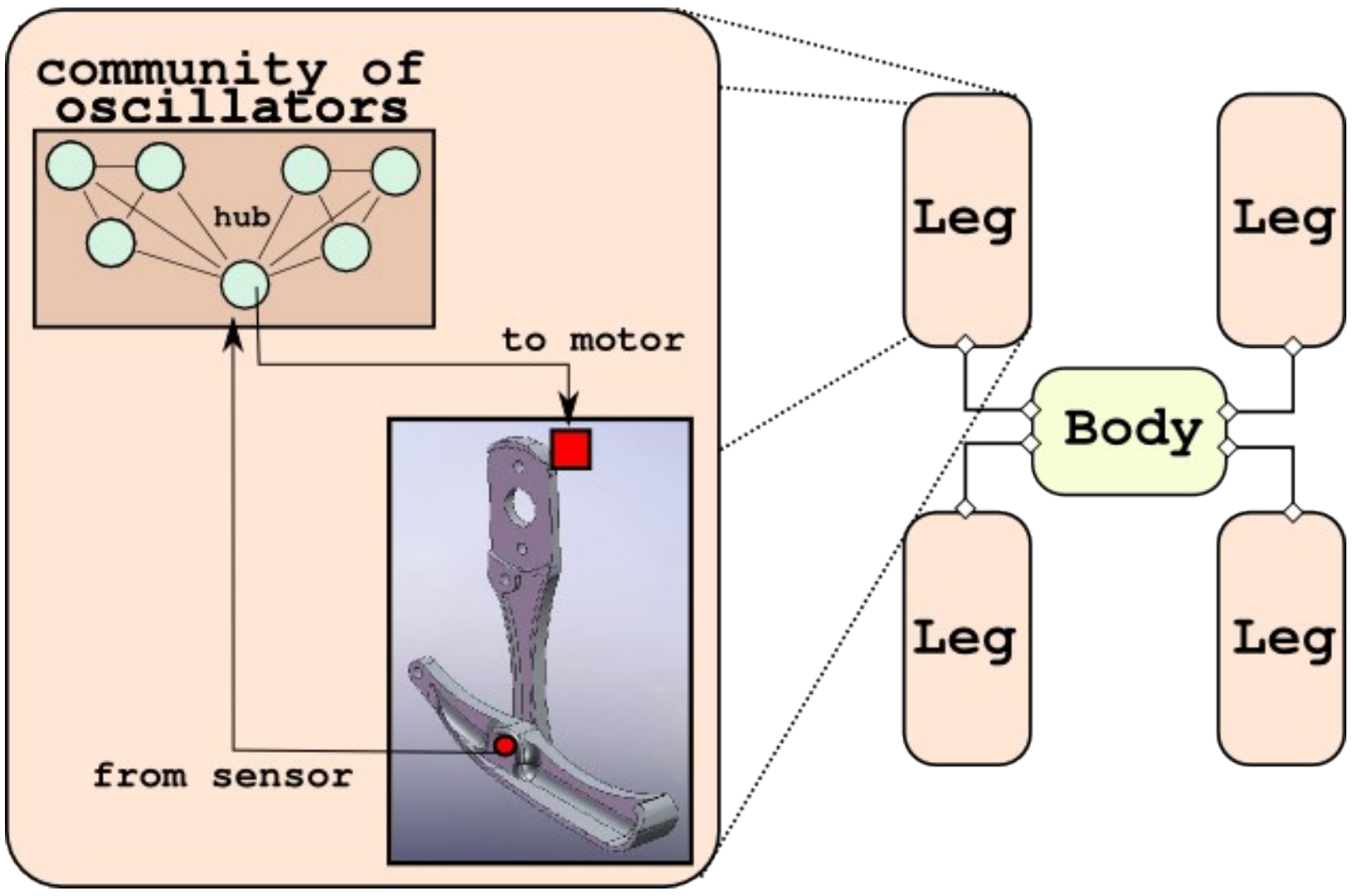

**Figure 16: Schematic of the proposed control architecture.** A CPG (central pattern generator) represented by a community of oscillators is influenced only by local information coming from the limb it controls. We are trying to determine whether this community can share information with the communities connected to the other limbs, through the environment and the body.

### *Controllers with chaotic dynamics*

Another strategy that we are currently employing that addresses the issue of exploration of movement possibilities are controllers featuring chaotic dynamics. However, a Master thesis (Hagmann, 2010) is currently at the stage of developing a suitable simulator.

### *Modeling – Normal modes*

Our work is centered on how the properties of the body facilitate behavior, particularly locomotion. The propelling idea is that coordinated motion patterns such as gaits (e.g. galloping, trotting, bounding etc. for for quadrupedal animals) are somehow natural to the morphology of the animal (or robot). To ground this notion into a verifiable theory and eventually into a design tool we proposed the use of the theory of *Normal modes* or *Structural dynamics*. The normal modes of a mechanical system are single frequency solutions to the equations of motion; the general motion of the system is a superposition of its normal modes. The modes are normal in the sense that they can be observed independently, the excitation of one mode will never excite a different mode. In many systems this is equivalent to reducing a collection of coupled oscillators to a set of decoupled, effective oscillators.

If we imagine a quadruped robot like Puppy with its feet attached to the ground (no flight phase), the similarities with the system shown in Figure 17 are clear. Accepting such a linear model, we can now ask what are the resonant modes of that structure and what actuation will cause them to be excited. In Figure 18 two example of possible normal modes are shown (depending on spring constant and plate dimensions). They are stotting (also pronking or pronging) and bounding gait.

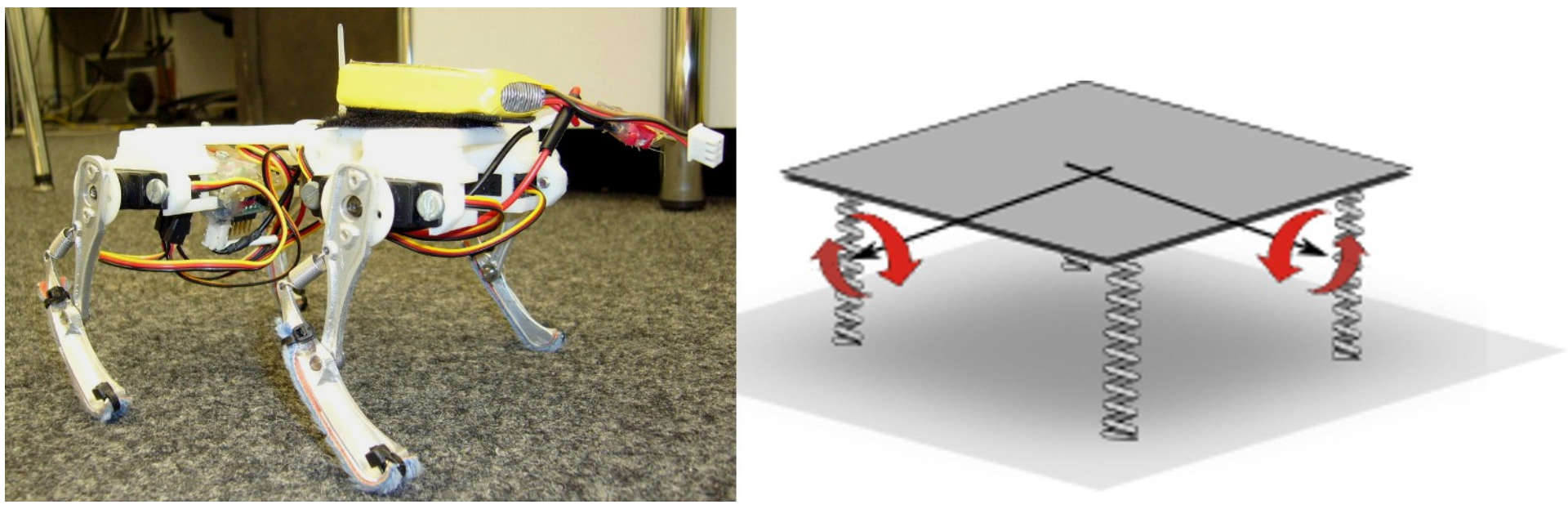

**Figure 17: Model of quadruped robot.** On the left a photo of the real robot. If we assume the legs are attached to the ground, it can be represented with the linear model shown on the right.

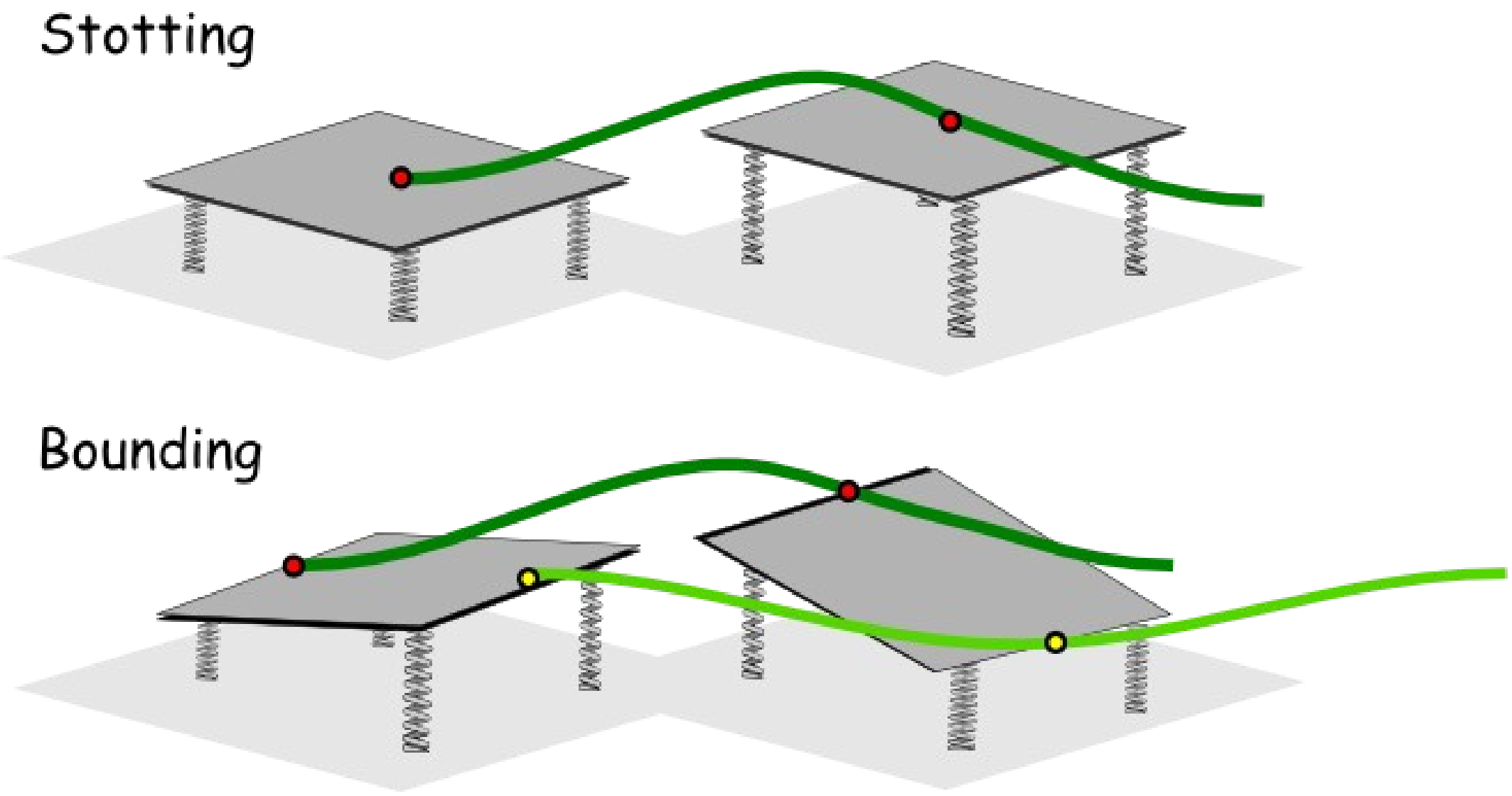

**Figure 18: Normal modes of a plate with springs.** Two possible normal modes resembling stotting and bounding gaits of quadrupeds. The lines depict the trajectory of the markers in the plate.

However, the real platform is not attached to the ground. If we allow the system to be detached form the ground, i.e. to have a flight phase, the standard methods for finding the normal modes have to be adapted. The system is now nonlinear and is considered a hybrid system. However, it is still periodic and the normal modes can be estimated. This estimation is the focus of our present research. How would a robot with the desired normal modes be easier to control? If a robot is built such that the gaits are close to its normal modes, what is left to know is how can these modes be excited. Once this knowledge is gained, the design of a controller setting the correct forcing is greatly simplified.

**Identification of a platform.** The first step was to see which simplified model would describe a robot the best. With this aim, we tested the Puppy robot shown in Figure 19. The robot was compressed from above in such a way that the four legs show the same compression. Once there, it was suddenly released and the data from the knee angle sensors was recorded (step response of the system). The results are shown in Figure 19. In solid line the best fit (in least mean squares sense) of a spring-damper system is shown. It turns out that the front and hind legs do not share the same frequency. Given that the morphology of the legs is similar and that the spring in all of them are the same, the effect can be due to uneven mass distribution and posture of the legs. If we neglect the posture effect, to fit the measurements the front to rear mass relation should be 0.6.

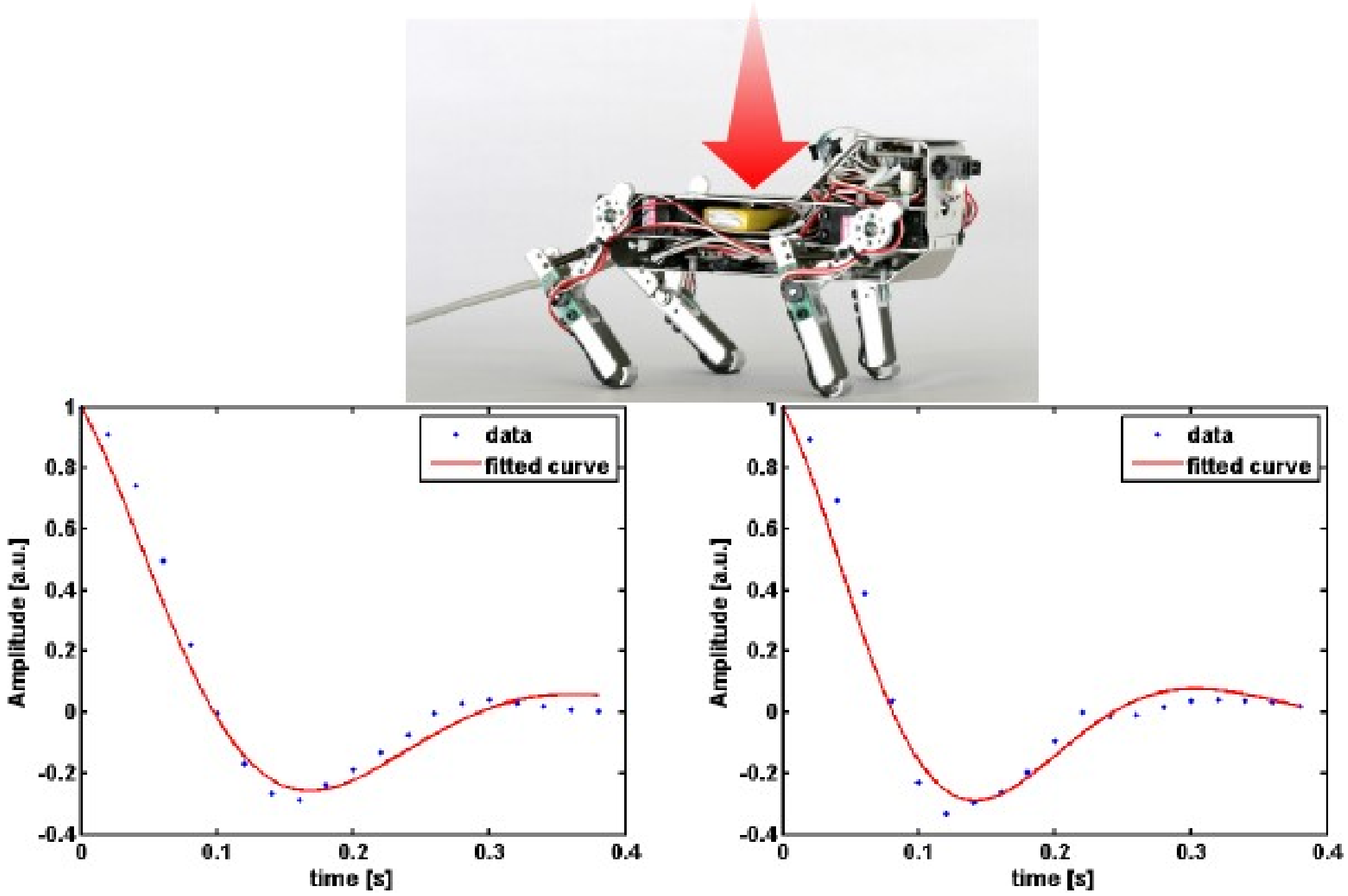

**Figure 19: Identification of Alan Puppy robot.** The step response of the legs of the robot are plotted. In solid line the best fit of a damper-spring system. It can be seen that the time scales of the responses differ; this can be associated with the effect of uneven mass distribution.

### Synthesis of body schema

Using our quadrupedal platforms, we are working on the synthesis or development of their body schema. Typically, by body schema a spatial representation of the robot's body in space is meant. However, we think that this view is too restricted. Therefore, for us body schema is rather that the robot can recognize its body, as well as its constraints and action possibilities. Also, while most research focuses on manual actions (e.g. how a hand representation in space arises from a combination of visual and tactile stimuli), we are dealing with a 'locomotor body schema'. For a quadrupedal robot, it is crucial that it knows the outcomes of its actions. One possible action is the use of a particular gait and the outcome is where this gait is going to bring the robot – a navigation problem. We are developing a model that allows the robot to navigate using only information from so-called self-motion cues, i.e. without an external reference system (such as visual landmarks in the animal kingdom, or GPS in typical robot applications) (Reinstein and Hoffmann, in preparation). The multimodal sensory information consists of accelerometers, angular rate sensors (gyroscopes), angular position sensors on joints, and pressure sensors on feet. Fusing these together, the robot can synthesize a navigation system that uses self-motion cues only and allows it to integrate its path from its 'nest', for instance. The key variables that it needs to derive are change in position and heading. We are constructing a model that will fuse information from two sources: (i) an inertial navigation system (which uses accelerometers and angular rate sensors); and (ii) 'virtual odometer'. The latter is using joint angle information and pressure sensors. The fusion of information is accomplished by means of a Kalman filter that is estimating the errors of the two sources (Fig. 20 (b)).

The virtual odometer relies on the fusion of sensors other than those from the inertial measurement unit, as these are already exploited by the inertial navigation system. Speed and change in heading need to be derived. Speed can in turn be obtained as a product of motor frequency (which is known) and stride length (which is unknown). To construct the virtual odometer, we are looking for indicators based on angular position sensors on joints and pressure sensors on feet and their correlations with stride length and change in heading (Fig. 21). We are addressing different gaits, speeds, and terrains with both simulated and real robot.

(b)

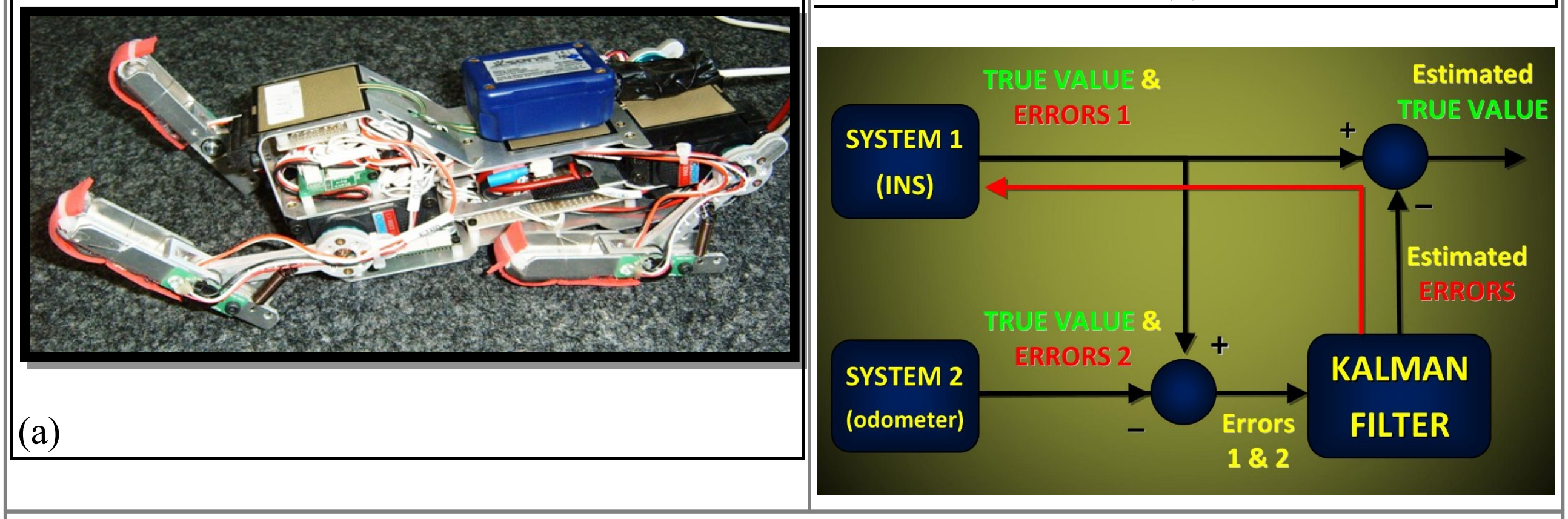

(a)

**Figure 20: Self-motion cue based navigation in the quadrupedal robot.** (a) Puppy II robot equipped with an additional Inertial Measurement Unit (blue box). (b) Fusion and error estimation algorithm. Information about the robot's position, velocity and attitude is fused and errors are estimated and subtracted with a Kalman filter.

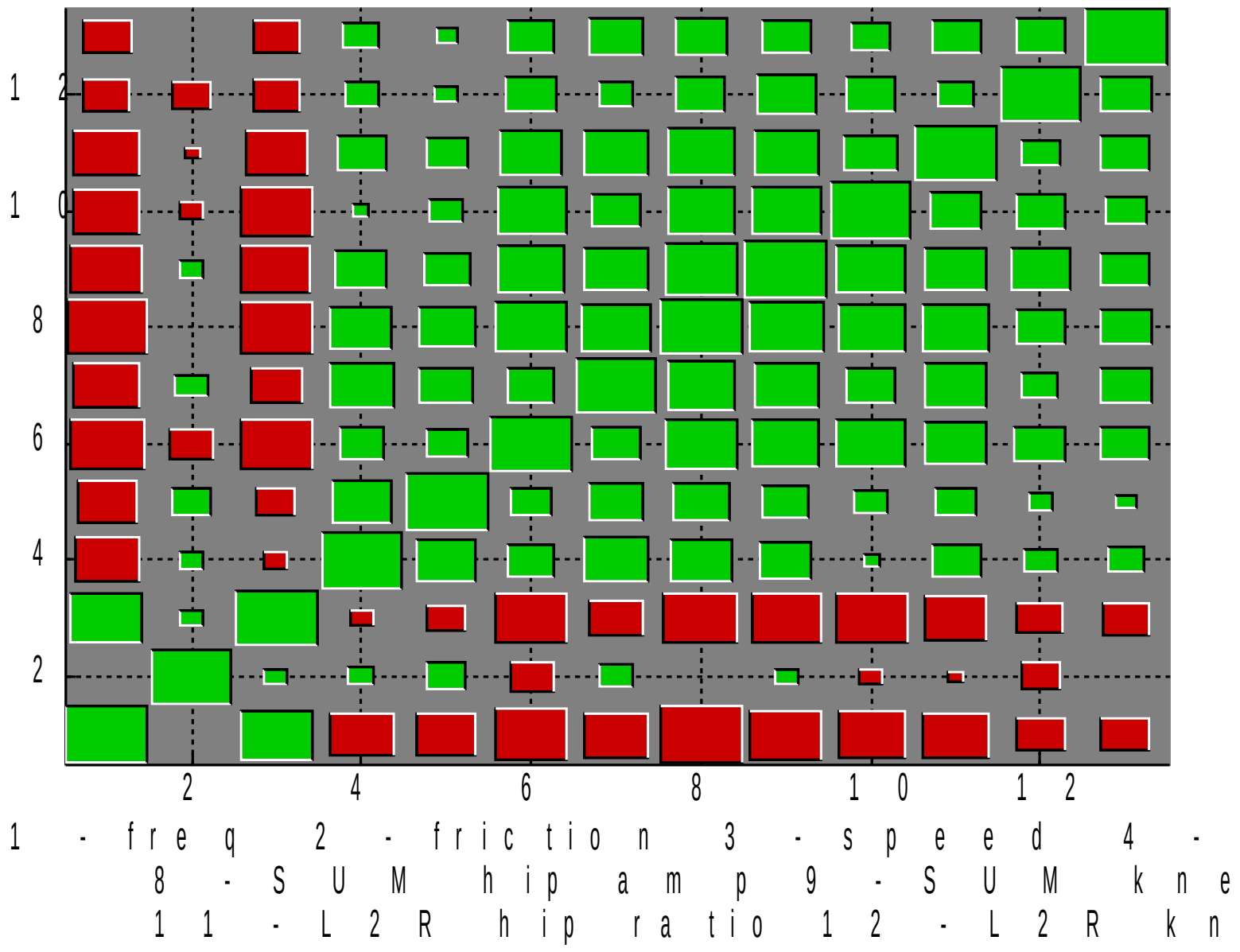

**Figure 21: Correlation matrix for virtual odometer.** Data comes from the Webots simulator where a simulated quadruped was run with a turning gait, on terrains with different friction. Columns 4 and 5 are the variables of interest (stride length and delta heading). Green squares represent positive correlations, red squares negative, and the size of square is proportional to the correlation size. The other columns represent potential indicators derived from other sensors. For instance, a sum of hip amplitudes (column 8) can be seen to correlate positively with stride length.

As the next stage, the robot can use its 'locomotor body schema' to plan trajectories. To test this, we have come up with a predator-prey scenario, where a (so far simulated) quadruped robot is 'hunting' another quadruped robot. A model of individual gaits and their transitions is necessary. This is work in progress and will constitute and natural transition to Work package 3 that we will be working on in the next project period.

# 2 Overview of contributions of SNF researchers

## Funded by the project

Marc Ziegler – Underwater locomotion.
Matej Hoffmann – Quadrupedal locomotion.
Juan Pablo Carbajal – Legged and swimming platforms – systematic exploration of morphology.

## Supervision of MSc., BSc. theses

Nuesch, S. (2009). Hopf oscillator with sensory feedback for adaptive robot locomotion. Unpublished Master thesis, University of Zurich.

Michel, M. M. (2009). Synchronization of dynamical systems for gait emergence in quadruped robots. Unpublished Master thesis, Department of Information Technology and Electrical Engineering (D-ITET), ETH Zurich, and University of Zurich.

Hutter, S. (2009). Co-evolution of morphology and controller of a simulated underactuated quadruped robot using evolutionary algorithms. Unpublished Bachelor thesis, University of Zurich.

Faessler, U., and Ruegg, N. (2009). A robot learning to walk. Unpublished Bachelor thesis, ZHAW School of Engineering, and University of Zurich.

Benker, E. (2009). Tunable springs. Unpublished Bachelor thesis, Fachhochschule Nordwestschweiz (FHNW), and University of Zurich.

### *Ongoing theses*

Hagmann, E. (2010). Exploration of Body Capabilities through Feedback Resonance of Chaos. Master Thesis. ETH Zurich, and University of Zurich.

# 3 Dissemination and special events

### *«SCIENCEsuisse» portrait of Rolf Pfeifer (November 2008)*

A documentary about Rolf Pfeifer in the context of the "SCIENCEsuisse" series had its premiere in November 2009. The series features 25 leading Swiss researchers. The documentary was entitled "Intelligence of the body" and was featuring the robots and concepts developed in the context of this project. More information can be found here: http://www-internet.sf.tv/sendungen/sciencesuisse/manualx.php?docid=rolf-pfeifer.

### *Lecture at Kinderuniversität (19. 11. 2008)*

Rolf Pfeifer gave a lecture at the 'Children University' with the title: Will robots soon be like humans? He was assisted by Matej Hoffmann who presented a quadruped robot demo. The audience consisted of approximately 500 children. More information can be found here: http://www.kinderuniversitaet.uzh.ch/HS_08/programm_HS08.html#roboter.

### *FET Conference Prague (21. - 23. 4. 2009)*

Matej Hoffmann and Juan Pablo Carbajal presented a poster of "From locomotion to cognition" at the European Future Technologies Conference, Prague. Link: http://ec.europa.eu/information_society/events/fet/2009/.

*Shanghai Science and Art Exhibition (14. - 20. 5. 2009)*

Rolf Pfeifer, Matej Hoffmann, and Pascal Kaufmann presented the research done at the AI Lab at the Shanghai Science and Art Exhibition. Two of the quadrupedal robots developed under this project were presented to numerous visitors in live demos. Our presence has received significant media coverage. The AI lab booth was part of the Swiss pavillon that has been awarded the "Creativity and Innovation Award."

*Keynote presentation 5th anniversary of Swissnex Singapore (06. 07. 2009) & Opening of SCIENCESuisse exhibition (07. - 28. 07. 2009)*

On the occasion of the $5^{th}$ anniversary of Swissnex Singapore, Rolf Pfeifer presented the AI lab and also the quadrupedal robot developed under this project. The robot was also presented at the opening of SCIENCESuisse exhibition in Singapore, Fusionopolis.

*Lab tours*

The project was presented to visitors (teachers, grammar school and high school students, representatives from companies, managers, staff from universities of applied science, etc.) in numerous lab tours (around 30).

## Lectures and invited talks

### Prof. Dr. Rolf Pfeifer

On the role of embodiment in enactive behavior (tentative title). Invited keynote lecture at the "Enactive 2008 Conference", Pisa, November 2008.

Self-organization, embodiment, and biologically inspired robotics. Invited keynote lecture. Darwin Days, Oslo, February 2009.

Artificial Intelligence. Samstagsuniversität, Bern, February 2009.

Workshop on designing robots through exploitation of morphological and material properties. Nanyang Technological University, Singapore, February 2009.

Embodied intelligence. Scuola Superiore Sant'Anna, Pisa, Italy, February 2009.

Cognition -- the interaction of brain, body, and environment. Invited keynote lecture. Third International Conference on Cognitive Science, Tehran, March 2009.

Embodied intelligence. FET 2009 Conference, Science Beyond Fiction, Prague, 2009.

Bodily intelligent modular robots. FET 2009 Conference, Science Beyond Fiction, 2009.

Self-organization, embodiment, and biologically inspired robotics. Keynote lecture, Lausanne, Switzerland, EPFL, Robotics Research Day, April 2009.

The four messages of embodied intelligence. Swissnex Fifth Anniversary Celebration, Singapore, July 2009.

Self-organization, embodiment, and biologically inspired robotics. Dept. of Automation, Jiao Tong University, Shanghai, July 2009.

Exploiting biomechanical constraints in rehabilitation robotics. Jiao Tong University, Shanghai, Chinese-Japanese-Singapore workshop on rehabilitation robotics. September 2009.

### Marc Ziegler

"Cheap" Underwater Locomotion. Invited talk at FILOSE Workshop on biomimetics, July 2009, Tallinn University of Technology, Estonia.

**Matej Hoffmann**

Embodied AI: The road of a robotic dog to cognition. Invited talk at Artificial beings seminar, Charles University, Prague, December 2008.

**Juan Pablo Carbajal**

New AI in robotics. Invited speaker (on behalf of Rolf Pfeifer). The Ninth Annual Meeting of EURON, Leuven, April 2009.

Artificial Intelligence and Robotics. New ideas, new opportunities. Open talk at the National University of Salta, Argentina, June 2009.

What is Artificial Intelligence? What should I study? Divulgation talk for high school students. Institute of secondary education (IEM), Argentina, June 2009.

# 4 Publication list

## Book and Journal

Hoffmann, M. & Pfeifer, R. (2009), 'Let animats live!' Adaptive behavior 17 (4), 317-319.

Pfeifer, R., and Gomez, G. (in press). Intelligence, the interaction of brain, body and environment. Design principles for adaptive systems. In S. Nefti-Meziani (ed.). Advances in Cognitive Systems.

Pfeifer, R. and Gómez, G. (2009). Morphological computation - connecting brain, body, and environment. In Körner, E., Sendhoff, B., Sporns, O., Ritter, H., and Doya, K. (Eds.) *Creating Brain-like Intelligence: Challenges and Achievements.* Springer-Verlag, Berlin, 66-83.

Pfeifer, R., Lungarella, M. and Sporns, O. (2008). The synthetic approach to embodied cognition: a primer. In: P. Calvo and T. Gomilla (eds.) Handbook of Cognitive Science.

## Publications in preparation

Reinstein, M. and Hoffmann, M. (2010). Dead reckoning in a legged robot.

Ziegler, M., Carbajal, J.P., and Pfeifer, R. (2010). Roles of resonance in underactuated robot swimming.

Ziegler, M., Hoffmann, M., and Pfeifer, R. (2010). Design of a controller for a tunable flexible tail fin in a robot fish.

## MSc. and BSc. theses

Nuesch, S. (2009). Hopf oscillator with sensory feedback for adaptive robot locomotion. Unpublished Master thesis, University of Zurich.

Michel, M. M. (2009). Synchronization of dynamical systems for gait emergence in quadruped robots. Unpublished Master thesis, Department of Information Technology and Electrical Engineering (D-ITET), ETH Zurich, and University of Zurich.

Hutter, S. (2009). Co-evolution of morphology and controller of a simulated underactuated quadruped robot using evolutionary algorithms. Unpublished Bachelor thesis, University of Zurich.

Faessler, U., and Ruegg, N. (2009). A robot learning to walk. Unpublished Bachelor thesis, ZHAW School of Engineering, and University of Zurich.

Benker, E. (2009). Tunable springs. Unpublished Bachelor thesis, Fachhochschule Nordwestschweiz (FHNW), and University of Zurich.